\documentclass[10pt,twocolumn,letterpaper]{article}

\usepackage{cvpr}              % To produce the CAMERA-READY version
\usepackage{graphicx}
\usepackage{amsmath}
\usepackage{amssymb}
\usepackage{booktabs}
\usepackage{color}
\usepackage{multirow}
\usepackage{tabularx}
\usepackage{caption}
\usepackage{silence}
\usepackage{hyperref}
\hypersetup{breaklinks,colorlinks}

\usepackage[capitalize]{cleveref}
\crefname{section}{Sec.}{Secs.}
\Crefname{section}{Section}{Sections}
\Crefname{table}{Table}{Tables}
\crefname{table}{Tab.}{Tabs.}

\def\cvprPaperID{5533} % *** Enter the CVPR Paper ID here
\def\confName{CVPR}
\def\confYear{2023}

\begin{document}

\title{The Euclidean Space is Evil: Hyperbolic Attribute Editing\\ for Few-shot Image Generation}

% \author{Lingxiao Li\\
% Columbia University\\
% \and
\author{
Lingxiao Li\textsuperscript{\textnormal{1}} \quad Yi Zhang\textsuperscript{\textnormal{2}} \quad Shuhui Wang\textsuperscript{\textnormal{3}}\thanks{Corresponding author}\\ \\\textsuperscript{1} Columbia University \quad
\textsuperscript{2} University of Oxford\\
\textsuperscript{3} Institute of Computing Technology, Chinese Academy of Sciences\\
{\tt\small ll3504@columbia.edu, wolf5965@ox.ac.uk, wangshuhui@ict.ac.cn}}

\maketitle

%%%%%%%%% ABSTRACT
\begin{abstract}
   Few-shot image generation is a challenging task since it aims to generate diverse new images for an unseen category with only a few images. Existing methods suffer from the trade-off between the quality and diversity of generated images. To tackle this problem, we propose Hyperbolic Attribute Editing~(HAE), a simple yet effective method. Unlike other methods that work in Euclidean space, HAE captures the hierarchy among images using data from seen categories in hyperbolic space. Given a well-trained HAE, images of unseen categories can be generated by moving the latent code of a given image toward any  meaningful directions in the Poincaré disk with a fixing radius. Most importantly, the hyperbolic space allows us to control the semantic diversity of the generated images by setting different radii in the disk. Extensive experiments and visualizations demonstrate that HAE is capable of not only generating images with promising quality and diversity using limited data but achieving a highly controllable and interpretable editing process. Code is available at \href{https://github.com/lingxiao-li/HAE}{https://github.com/lingxiao-li/HAE}.
\end{abstract}

%%%%%%%%% BODY TEXT
\section{Introduction}
\label{sec:intro}

Due to the persistent development of deep learning, the task of image generation has received signiﬁcant research attention in recent years.
Specifically, the Generative Adversarial Networks (GANs)~\cite{Goodfellow14} and its variants (\textit{e.g.}, StyleGANv2~\cite{Karras20}) have succeeded in generating high-fidelity and realistic images, requiring a large number of high-quality data for model training.
However, considering the long-tail distribution and data imbalance
widely exists among different image categories~\cite{Hong20F2}, it is difficult for GANs to be trained on categories with sufficient training images to generate new realistic and diverse images for a category with only a few images. 
This task is referred to as few-shot image generation~\cite{Clouâtre19, Hong20, Hong20F2, Hong20Match, Hong22, Hong22Delta, Ding22}. A variety of tasks can benefit from improvements in few-shot image generation, for instance, low-data detection~\cite{Fu19} and few-shot classification~\cite{Sung18, Vinyals16}.

\begin{figure}[t]
  \centering
  \setlength{\abovecaptionskip}{0.2cm}   %调整图片标题与图距离
  \setlength{\belowcaptionskip}{-0.2cm}   %调整图片标题与下文距离
   \includegraphics[width=1\linewidth]{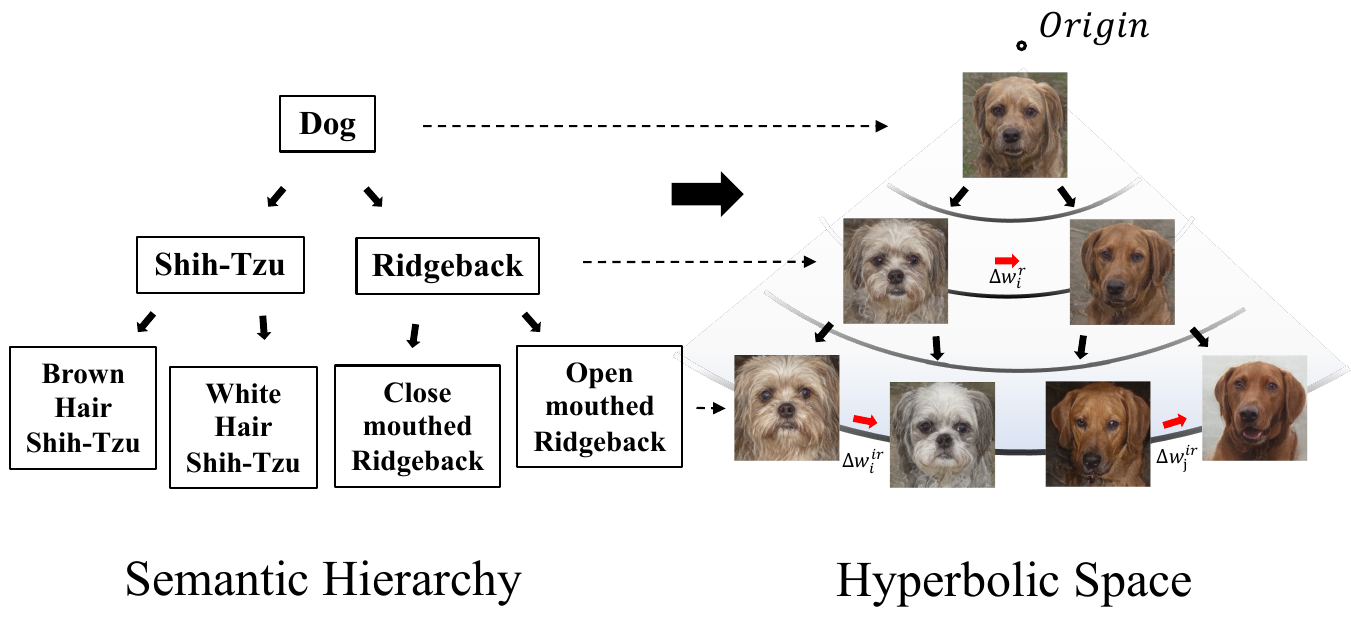}

   \caption{\textbf{Illustration of hierarchical attribute editing in hyperbolic space.} Hyperbolic space can naturally and compactly encode semantic hierarchical structures within a large image data corpus. Changing the high-level, \ie, category-relevant attribute $\Delta \mathbf{w}^{r}$ changes the category of an image. While changing low-level or category-irrelevant attribute $\Delta \mathbf{w}^{ir}$ varies images within categories.}
   \label{fig:1}
\end{figure}

In general, existing GAN-based few-shot image generation mechanisms can be classified into three categories. Transfer-based methods~\cite{Clouâtre19, Liang20} introduce meta-learning or domain adaptation on GANs to generate new images by enforcing knowledge transfer among categories. Fusion-based methods~\cite{Bartunov18, Gu21, Hong20F2, Hong20Match} perform feature fusion of multiple input images in a feature space and generate images via decoding the fused features back to image space. However, the output is still highly similar to the source images. Transformation-based methods~\cite{Antoniou17, Hong22, Hong22Delta, Ding22} find intra-category transformations or inject random perturbations to conditional unseen category samples to generate images without tedious fine-tuning. 
By representing the images in the Euclidean feature space, the above learning mechanisms tend to be over-complicated, and the generated images are often collapsed due to limited diversity.

%We observe that existing GAN inversion methods all work in Euclidean space, where common distance metrics can not handle hierarchical representation well.

Similar to the ubiquity of hierarchies in language~\cite{Nickel17, Tifrea19, Dhingra18}, the semantic hierarchy is also common in images~\cite{Khrulkov21, Cui23}. As \cref{fig:1} shows, the semantic hierarchies constructed in the language domain can be instantiated with visual images. From the visual perspective, an image can be regarded as a collection of attributes of multiple levels. High-level attributes, {\it a.k.a.} category-relevant attributes, define the category of an image, such as the shape and color of an animal~\cite{Ding22}. For instance, in the middle row of \cref{fig:1}, changing the high-level attributes of the given image of a \texttt{Shih-Tzu dog}, the category can be changed to a \texttt{Rhodesian Ridgeback Dog}. While the low-level or fine-grained attributes, including expressions, postures, \etc, that vary within the category as shown at the bottom of \cref{fig:1}, are called category-irrelevant attributes. Therefore, an image can also be viewed as a descendant of another image with the same category-relevant attributes by adding fine-grained category-irrelevant attributes to its parent image. 
To edit the visual attributes for high-quality image generation, 
it is crucial to capture the attribute \textit{hierarchy} within the large image data corpus and find a good representation space. Ideally, we aim to construct a hierarchical visual representation in a latent space that allows us to change the category of an image by moving the latent code in a category-relevant direction, and perform few-shot image generation by moving the code in a category-irrelevant direction.

Unfortunately, the Euclidean space and its corresponding distance metrics used by
existing GAN-based methods can not facilitate the hierarchical attribute representation, thus the design of complicated attribute disentangling and editing mechanisms seems to be crucial for the generation quality. Inspired by the application of hyperbolic space in images~\cite{Khrulkov21} and videos~\cite{Surís21}, we found that the metrics introduced in hyperbolic geometry can naturally and compactly encode hierarchical structures. Unlike the general affine spaces, \eg, the Euclidean space, hyperbolic spaces can be viewed as the continuous analog of a tree since tree-like graphs can be embedded in ﬁnite-dimension with minimal distortion~\cite{Nickel17}. This property of hyperbolic space provides continuous and up to infinite semantic levels for attribute editing, allowing us to robustly generate diverse images with only a few images from unseen categories with simple operations.

Based on the above findings, we propose a simple but effective Hyperbolic Attribute Editing (HAE) method for few-shot image generation. Our method is based on the observation that hierarchical latent code manipulation can be easily implemented in Hyperbolic space. The core of HAE is mapping the latent vectors from the Euclidean space $\mathbb{R}^n$ to a hyperbolic space $\mathbb{D}^n$. We minimize a supervised classification loss function to ensure the images are hierarchically embedded in hyperbolic space. Once we capture the attribute hierarchy among images, we can generate new images of unseen categories by moving the latent code from one leaf to another with the same parents by fixing the radius. Most importantly, the hyperbolic space allows us to control the semantic diversity of generated images by setting different radii in the Poincaré disk. Those operations can well facilitate continually hierarchical attribute editing in hyperbolic space for flexible few-shot image generation with both quality and diversity.

Our contributions can be summarized as follows:
\begin{itemize}
    \item We propose a simple yet effective method for few-shot image generation,~\ie, hyperbolic attribute editing. In order to capture the hierarchy among images, we use hyperbolic space as the latent space. To the best of our knowledge, HAE is the first attempt to use hyperbolic latent spaces for few-shot image generation. 
    \item We show that in our designed hyperbolic latent space, the semantic hierarchical attribute relations among images can be reflected by their distances to the center of the Poincaré disk. 
    \item Extensive experiments and visualization suggest that HAE achieves stable few-shot image generation with state-of-the-art quality and diversity. Unlike other few-shot image generation methods, HAE allows us to generate images with better control of diversity by changing the semantic levels of attributes we want to edit.
\end{itemize}
%-------------------------------------------------------------------------
\begin{figure*}[t]
  \centering
  \setlength{\abovecaptionskip}{0.2cm}   %调整图片标题与图距离
  \setlength{\belowcaptionskip}{-0.2cm}   %调整图片标题与下文距离
   \includegraphics[width=1\linewidth]{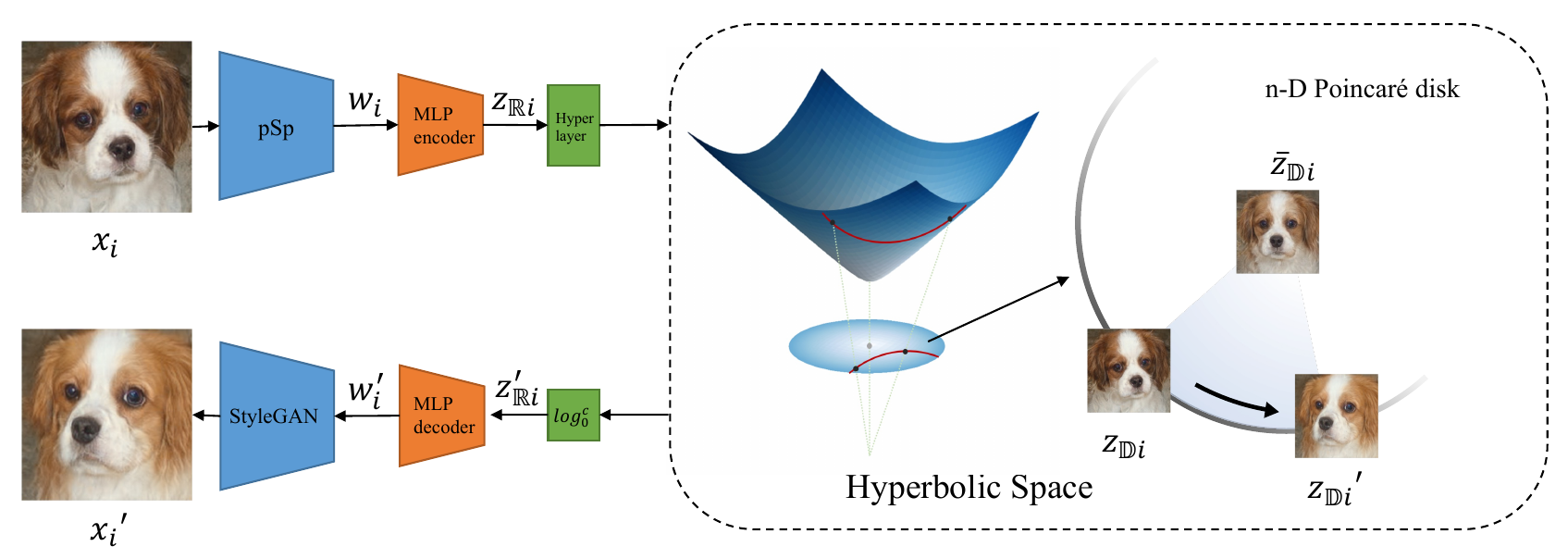}

   \caption{The overview of HAE. The \textbf{Hyper layer} is a hyperbolic feedforward layer called \textit{Möbius linear layer} which is used to project the latent code from Euclidean space $\mathbb{R}^n$ to hyperbolic space $\mathbb{D}^n$~\cite{Chami19}. 
   %The hierarchical representation in hyperbolic space allows us to perform hierarchical attribute editing. 
   $\bar{z}_{\mathbb{D}i}$ can be viewed as the ``parent" or average code of $z_{\mathbb{D}i}$ and $z_{\mathbb{D}i}'$. One can generate diverse images without changing the category by moving the latent code from one child to another of the same parent in the hyperbolic space.}
   \label{fig:flowchart}
\end{figure*}

\section{Related Work}

\noindent \textbf{Few-shot image generation}. Recently, diverse methods have been proposed for few-shot image generation. The transfer-based methods~\cite{Clouâtre19, Liang20} which introduce meta-learning or domain adaptation on GANs can hardly generate realistic images. While fusion-based methods that fuse the features by matching the random vector with the conditional images~\cite{Hong20Match} or formulating the problem as a conditional generating task~\cite{Gu21, Hong20F2} suffer from the limited diversity of generated images. Furthermore, transformation-based methods~\cite{Antoniou17, Hong22, Hong22Delta, Ding22} can generate images with only one conditional image by focusing on either capturing the cross-category or intra-category transformations by injecting random perturbations ~\cite{Antoniou17}. Nevertheless, the transformation captured by those methods is not very consistent. Ding {\it et al.}~\cite{Ding22, Ding23} propose the ``editing-based” perspective, the intra-category transformation can be modeled as category-irrelevant image editing based on one sample instead of pairs of samples. Most recently, Zhu {\it et al.}~\cite{Zhu22}  fine-tune powerful diffusion models (DMs)~\cite{Ho20} pre-trained on large source domains on limited target data to generate diverse and high quality images. DMs outperform GANs~\cite{Goodfellow14} on sample quality with a more controllable training process at the cost less flexibility and editability, since they denoise images in the image space rather than operate in the latent space. Furthermore, the inference process of DMs is much slower than GANs~\cite{Song21DDIM}.

\noindent \textbf{Hyperbolic Embedding}. The use of hyperbolic space in deep learning~\cite{Nickel17,Nickel18,Tifrea19,Surís21,Khrulkov21} is a pioneering work in recent years. It was first used in natural language processing for hierarchical language representation~\cite{Nickel17,Nickel18,Tifrea19}. The Riemannian optimization algorithms are used to optimize models in hyperbolic space~\cite{Bécigneul19,Bonnabel13}. As hyperbolic space is successfully applied to represent hierarchical data, Ganea {\it et al.}~\cite{Ganea18} derives hyperbolic versions of tools in neural networks including multinomial logistic regression, feed-forward, and recurrent neural networks. Following this, hyperbolic geometry is used in image~\cite{Khrulkov21}, video~\cite{Surís21}, and graph data~\cite{Chami19, Park21}. Most recently, Lazcano {\it et al}~\cite{Lazcano21} shows that hyperbolic space outperforms traditional Euclidean space in image generation using HGAN. However, the hierarchy and controllability of hyperbolic space remain uninvestigated in HGAN, as the generator is still governed by Gaussian samples in Euclidean space.

\noindent \textbf{Latent Code Manipulation}. It has been shown that the latent spaces of GANs are able to encode rich semantic information~\cite{Goetschalckx19,Jahanian20,Shen20}. One of the popular approaches is finding linear directions corresponding to changes in a given binary labeled attributes, which might be difﬁcult to obtain for new datasets and could require manual labeling effort~\cite{Shen20,Goetschalckx19,Denton19}. Others~\cite{Cherepkov21,Voynov20,Lu20,Härkönen20,Choi22} try to find semantic directions in an unsupervised manner. For instance, PCA is applied in the latent space to create interpretable controls for synthesizing images~\cite{Härkönen20,Choi22}. Most recent works~\cite{Eliezer21,Shen21} directly compute in the close form to find the meaningful semantic direction without training and optimization. In comparison, our work HAE focuses on attributes in different semantic levels in the latent space rather than trying hard to find disentangled interpretable directions as previous works. 

\section{Method}
The overall framework of HAE is shown in \cref{fig:flowchart}, we first give a detailed explanation of getting the hierarchical representations in the hyperbolic space, and then we introduce the framework of HAE and explain the loss functions.

\subsection{Hierarchical Representation}
\label{sec:Hierarchical Representation}
The major issue of our study is how to obtain the hierarchical representation from real images to facilitate editing in different semantic levels, as illustrated in \cref{fig:1}. 
%For instance, in \cref{fig:1}, we want to change the color and shape (category-relevant feature) of the Spaniel to change its category to meerkat. At the same time, we also want to change the posture or light (category-irrelevant) of the given image of the Spaniel to make diverse images of the Spaniel. 
Therefore, \textit{hyperbolic} space is introduced as the latent space to achieve this goal. 

Unlike Euclidean spaces with their zero curvature and spherical spaces with their positive curvature, hyperbolic spaces with negative curvature have been shown that it is more appropriate for learning hierarchical representation~\cite{Nickel17,Nickel18}. Informally, hyperbolic space can be viewed as a continuous analogy of trees~\cite{Nickel17}. One important feature of hyperbolic space is that the length grows exponentially with its radius while linearly in Euclidean space. This property allows hyperbolic space to be naturally compatible with hierarchical data~\cite{Gromov87} including text, images, videos, \etc.

The $n$-dimensional hyperbolic space can be formally defined as a homogeneous, simply connected $n$-dimensional Riemannian manifold, denoted as $\mathbb{H}^n$ with constant negative sectional curvature\footnote{The curvature of the hyperbolic space $c$ is set as $-1$ in this work.}. We choose to work in the Poincaré disk from ﬁve isometric models of hyperbolic space defined in~\cite{Cannon97} since it is commonly used in gradient-based learning~\cite{Nickel17,Ganea18,Nickel18,Tifrea19,Surís21,Khrulkov21}. The Poincaré disk model $\left(\mathbb{D}^n, g^{\mathbb{D}}\right)$ is defined by the manifold $\mathbb{D}^n=\left\{x \in \mathbb{R}^n:\|x\|<1\right\}$ equipped with the following Riemannian metric:
\begin{equation}
    g_x^{\mathbb{D}}=\lambda_x^2 g^E,
    \label{eq:Riemannian metric}
    \end{equation}
where $\lambda_x =\frac{2}{1-\|x\|^2}$, and $g^E$ is the Euclidean metric tensor $g^E = \mathbf{I}^n$. The induced distance between two points $\mathbf{x}, \mathbf{y} \in \mathbb{D}^n$ can be defined by:
\begin{equation}
    d_{\mathbb{D}}(\mathbf{x}, \mathbf{y})=\mathrm{arccosh}\left(1+2 \frac{\|\mathbf{x}-\mathbf{y}\|^2}{\left(1-\|\mathbf{x}\|^2\right)\left(1-\|\mathbf{y}\|^2\right)}\right).
    \label{eq3-distance}
\end{equation}

\begin{figure}[t]
  \centering
  \setlength{\abovecaptionskip}{0.2cm}   %调整图片标题与图距离
  \setlength{\belowcaptionskip}{-0.2cm}   %调整图片标题与下文距离
   \includegraphics[width=0.9\linewidth]{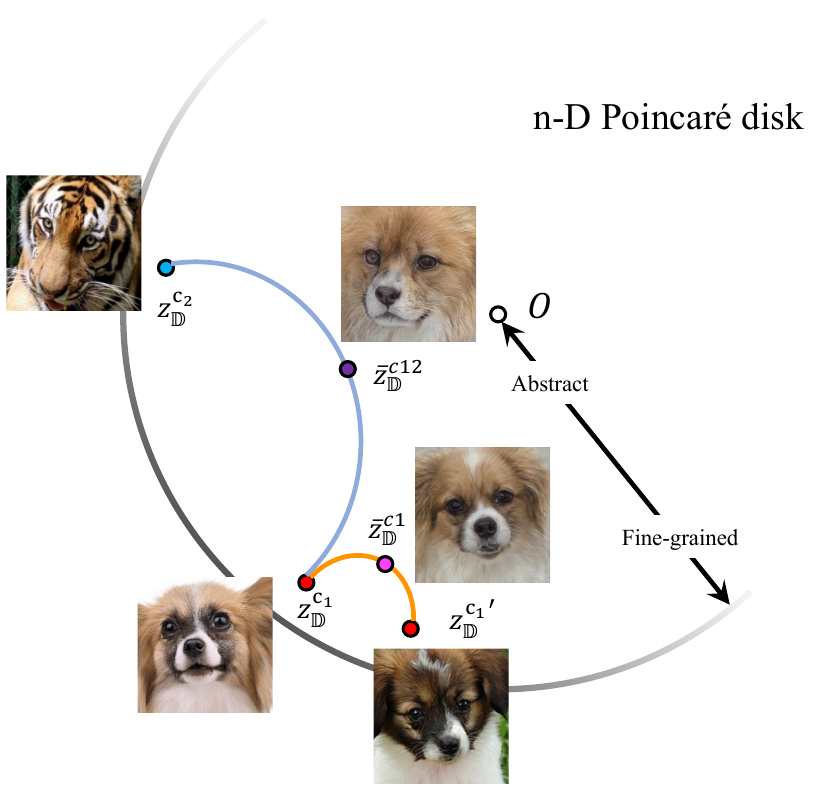}

   \caption{\textbf{Illustration of the property of hyperbolic space on the Poincaré disk.} Given two latent codes of Spaniel $z_{\mathbb{D}}^{c_1}$ and ${z_{\mathbb{D}}^{c_1}}'$(\textcolor{red}{red} dots) on the edge of Poincaré disk, the geodesic between these two points is the \textcolor[rgb]{0.588, 0.294, 0}{brown} curve rather than a straight line in Euclidean space. Therefore, their average latent code is calculated as $\Bar{z}_{\mathbb{D}}^{c_1}$(\textcolor{magenta}{pink} dot) which is closer to the center $O$ (still a Spaniel, but less fine-grained). While the latent code of a tiger $z_{\mathbb{D}}^{c_2}$(\textcolor{cyan}{blue} dot) locates far from the latent code of a Spaniel. Thus, the hyperbolic average code of tiger  and Spaniel $\Bar{z}_{\mathbb{D}}^{c_{12}}$(\textcolor[rgb]{0.627, 0.125, 0.941}{purple} dot) is closer to the center $O$ than $\Bar{z}_{\mathbb{D}}^{c_1}$ which is more abstract (a feline contains features from both tiger and Spaniel).} 
   \label{fig:disk}
\end{figure}

Recall that a geodesic is a locally minimized-length curve between two points. In the hyperboloid model, the geodesic can be defined as the curve created by intersecting the plane defined by two points and the origin with the hyperboloid~\cite{Lee13}. Thus, the mean of two latent codes in hyperbolic space locates at the mid-point of the geodesic that is closer to the origin. This is the key desired feature of hyperbolic space, \ie, the mean between two leaf embeddings is not another leaf embedding, but the hierarchical parent of them~\cite{Surís21}. This feature allows us to generate new images by moving the latent code from one leaf to another with the same parents. We can also change the semantic levels of attributes by determining how abstract their parent is.

This unique property is visualized in ~\cref{fig:disk} on a 2-D Poincaré disk. The image embedding near the edge of the ball (with a large radius) represents a more fine-grained image while the embedding near the center (which has a smaller radius) represents an image with abstract features (an average face). 

Although the hyperbolic space shares similar features with trees, it is continuous. In other words, there is no fixed number of hierarchy levels. Instead, there is a continuum from very fine-grained (near the edge of Poincaré disk) to very abstract (near the origin).

\subsection{Network Architecture}
Although we aim to embed and edit real images in hyperbolic space, the whole network does not need to be implemented in a hyperbolic manner. Instead, we can take advantage of the number of existing GAN inversion models and optimization algorithms that have been fine-tuned for Euclidean space. 

To achieve image editing, we need to embed the image back into the latent space. In particular, we select pSp~\cite{Richardson21} as the backbone of HAE to encode images to the $\mathcal{W^+}$-space of StyleGAN2~\cite{Karras20}:
\begin{equation}
   \mathbf{w}_i=\mathtt{pSp}\left(x_i\right), 
\end{equation}
where $\mathbf{w}_i \in \mathbb{R}^{18 \times 512}$ is the corresponding latent vector of $x_i$ in the $\mathcal{W^+}$-space.

To manipulate latent code in hyperbolic space, we need to define a bijective map from $\mathbb{R}^n$ to $\mathbb{D}_c^n$ to map Euclidean vectors to the hyperbolic space and vice versa. A manifold is a differentiable topological space that locally resembles the Euclidean space $\mathbb{R}^n$~\cite{Lee06,Lee13}. For $x \in \mathbb{D}^n$, one can define the tangent space $T_x \mathbb{D}^n_c$ of $\mathbb{D}^n_c$ at $x$ as the first order linear approximation of $\mathbb{D}^n_c$ around $x$. Therefore, this bijective map can be performed by exponential and logarithmic maps.
Specifically, the \textit{exponential map} $\exp _{\mathbf{x}}^c: T_{\mathbf{x}} \mathbb{D}_c^n \cong \mathbb{R}^n \to \mathbb{D}_c^n$, maps from the tangent spaces into the manifold. While the \textit{logarithmic map} $\log _{\mathbf{x}}^c: \mathbb{D}_c^n \to T_{\mathbf{x}} \mathbb{D}_c^n \cong \mathbb{R}^n$ is the reverse map of the exponential map.

\begin{figure*}[t]
  \centering
  \setlength{\abovecaptionskip}{0.2cm}   %调整图片标题与图距离
  \setlength{\belowcaptionskip}{-0.2cm}   %调整图片标题与下文距离
   \includegraphics[width=0.8\linewidth]{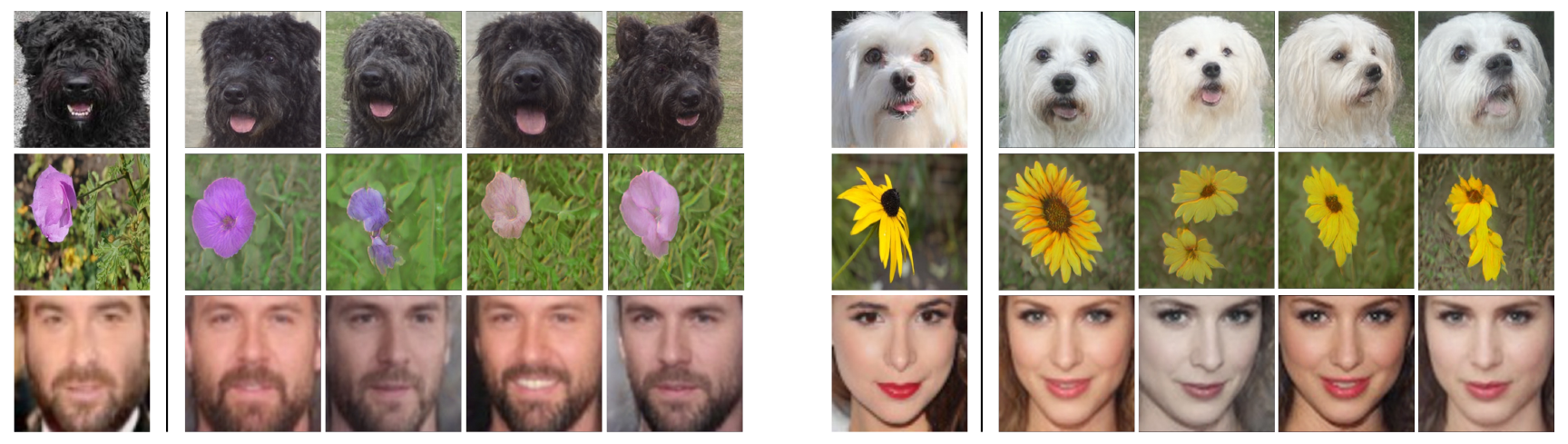}

   \caption{One-shot image generation from HAE on Animal Faces, Flowers, and VGGFaces.}
   \label{fig:generation}
\end{figure*}

We use exponential and logarithmic maps at origin $\textbf{0}$ for the transformation between the Euclidean and hyperbolic representations. After getting $\mathbf{w}_i$ in the $\mathcal{W^+}$-space, we first use a Multi-layer Perceptron (MLP) encoder to reduce the dimension of latent vectors in Euclidean space. Then we apply an exponential map to project the Euclidean latent code $z_{\mathbb{R}i}$ to hyperbolic space. After that, we use the hyperbolic feed-forward layer as ~\cite{Ganea18} to obtain the final hierarchical representation $z_{\mathbb{D}}$ as shown in Fig.~\ref{fig:flowchart}:
\begin{equation}
\label{eq:4}
    z_{\mathbb{D}i} = f^{\otimes_c}(\exp _{\mathbf{0}}^c(\mathtt{MLP}_E(\mathbf{w}_i))),
\end{equation}
where $f^{\otimes_c}$ is the Möbius translation of feed-forward layer $f$ as the map from $\mathbb{D}_c^n$ to $\mathbb{D}_c^m$, denoted as  \textit{Möbius linear layer}.

Finally, the hyperbolic representation $z_{\mathbb{D}}$ needs to be projected back to the $\mathcal{W^+}$-space of StyleGAN2. In practice, this is achieved by applying a logarithmic map followed by an MLP decoder:
\begin{equation}
\label{eq:5}
    \mathbf{w}_i' = \mathtt{MLP}_D(\log _{\mathbf{0}}^c(z_{\mathbb{D}i})),
\end{equation}
and $\mathbf{w}_i'$ will be fed into a pre-trained StyleGAN2's generator $G$ to reconstruct the image $x_i'$.

\subsection{Loss Function}
The loss function of HAE consists of two parts: the \textit{Hyperbolic loss} ensures to get the hierarchical representation in the hyperbolic space and the \textit{reconstruction loss} guarantees the quality of reconstruction images. 

\noindent \textbf{Hyperbolic Loss.} To learn the semantic hierarchical representation of real images in hyperbolic space, we minimize the distance between latent codes of images with similar categories and attributes while pushing away the latent codes from different categories. We choose the supervised approach to achieve this. In order to perform multi-class classiﬁcation on the Poincaré disk defined in \cref{sec:Hierarchical Representation}, one needs to generalize multinomial logistic regression (MLR) to the Poincaré disk defined in~\cite{Ganea18}. An extra linear layer needs to be trained for the classification and the softmax probability can be computed as:
Given $K$ classes and $k \in\{1, \ldots, K\}, p_k \in \mathbb{D}_c^n, a_k \in T_{p_k} \mathbb{D}_c^n \backslash\{\mathbf{0}\}$ :
\begin{equation}
\begin{aligned}
    p(y=k \mid x) \propto \exp &\Biggl(\frac{\lambda_{p_k}^c\left\|a_k\right\|}{\sqrt{c}} \sinh ^{-1}\\
    &\biggl(\frac{2 \sqrt{c}\left\langle-p_k \oplus_c x, a_k\right\rangle}{\left(1-c\left\|-p_k \oplus_c x\right\|^2\right)\left\|a_k\right\|}\biggl)\Biggl),\\ 
    & \quad \forall x \in \mathbb{D}_c^n.
\end{aligned}
\end{equation}

where $\oplus_c$ denotes the Möbius addition defined in ~\cite{Khrulkov21} with fixed sectional curvature of the space, denoted by $c$. 

After getting the softmax result for each class, one can use \textit{negative log-likelihood loss} (NLL Loss) to calculate the hyperbolic loss:
\begin{equation}
\label{eq:NLLLoss}
    \mathcal{L}_{\text {hyper}} = -\frac{1}{N}\sum_{n=1}^{N}\log(p_n),
\end{equation}
where $N$ is the batch size and $p_n$ is the probability predicted by the model for the correct class.

As mentioned in \cref{sec:Hierarchical Representation}, the distance between points grows exponentially with their radius in the Poincaré disk. In order to minimize \cref{eq:NLLLoss}, the latent codes of fine-grained images will be pushed to the edge of the ball to maximize the distances between different categories while the embedding of abstract images (images have common features from many categories) will be located near the center of the ball. 
Since hyperbolic space is continuous and differentiable, we are able to optimize \cref{eq:NLLLoss} with stochastic gradient descent, which learns the hierarchy of the images. 

\noindent \textbf{Reconstruction Loss.} In order to guarantee the quality of the generated images, we first use the $\mathcal{L}_2$ loss and LPIPS loss used in pSp~\cite{Richardson21}, given image $x_i$:
\begin{equation}
    \mathcal{L}_2(x_i)=\|x_i-\mathtt{HAE}(x_i)\|_2.
\end{equation}
\begin{equation}
    \mathcal{L}_{\text {LPIPS }}(x_i)=\|F(x_i)-F(\mathtt{HAE}(x_i))\|_2,
\end{equation}
where $F(\cdot)$ denotes the perceptual feature extractor.

Since the pSp encoder and StyleGAN2 generator are pre-trained, we only train the neural layers between the encoder and generator of $\mathtt{HAE}$. To further guarantee the network to better project back to the $\mathcal{W^+}$-space, the reconstructed $\mathbf{w}_i'$ should be the same as the original $\mathbf{w}_i$:
\begin{equation}
    \mathcal{L}_{\text{rec}}(w_i)=\|\mathbf{w_i}-\mathbf{w}_i'\|_2,
\end{equation}
where $\mathbf{w}_i'$ can be calculated by \cref{eq:4} and \cref{eq:5}.

The \textbf{overall loss function} is:
\begin{equation}
    \mathcal{L} = \mathcal{L}_2(x_i) + \lambda_1 \mathcal{L}_{\text {LPIPS}} + \lambda_2 \mathcal{L}_{\text{rec}} + \lambda_3 \mathcal{L}_{\text {hyper}},
\end{equation}
where $\lambda_1$, $\lambda_2$ and $\lambda_3$ are trade-off adaptive parameters. This curated set of loss functions ensures the model learns the hierarchical representation and reconstructs images.

\begin{figure*}[t]
  \centering
  \setlength{\abovecaptionskip}{0.2cm}   %调整图片标题与图距离
  \setlength{\belowcaptionskip}{-0.2cm}   %调整图片标题与下文距离
   \includegraphics[width=0.8\linewidth]{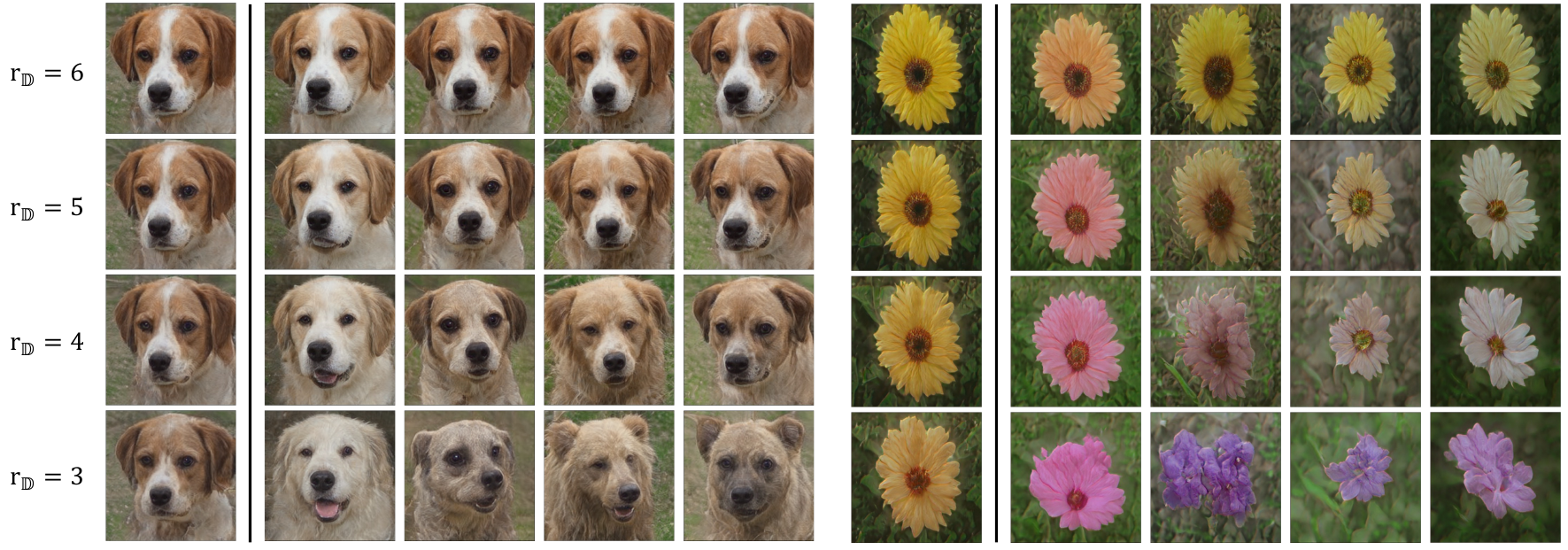}

   \caption{Images generated by HAE by adding the same perturbation on the latent code of a given image with different hyperbolic radii on Animal Faces and Flowers.}
   \label{fig:perturbation}
\end{figure*}

\subsection{Image Generation}
\label{Image Generation}
To study the generating quality of the model, a straightforward way is to generate new images via interpolation between two designated images, or random perturbation.

In hyperbolic space, the shortest path with the induced distance between two points is given by the geodesic defined in \cref{eq3-distance}. The geodesic equation between two embeddings $z_{\mathbb{D}\mathbf{i}}$ and $z_{\mathbb{D}\mathbf{j}}$, denoted by $\gamma_{z_{\mathbb{D}i} \rightarrow z_{\mathbb{D}j}}(t)$, is given by 
\begin{equation}
    \gamma_{z_{\mathbb{D}i} \rightarrow z_{\mathbb{D}j}}(t)= z_{\mathbb{D}i} \oplus_c t \otimes_c\left((-z_{\mathbb{D}i}) \oplus_c z_{\mathbb{D}j}\right), \: t \in [0,1],
\end{equation}
where $\oplus_c$ denotes the Möbius addition with aforementioned sectional curvature $c$, with details in supplementary material.

We adopt the following method to achieve generating via perturbation: For a given image $x_i$, we first rescale its embedding $z_{\mathbb{D}i}$ to the desired radius $r_\mathbb{D}$. Then we sample a random vector from seen categories in $z_{\mathbb{D}j}$ with radius $r_\mathbb{D}$ fixed and take the geodesic as the direction of perturbation to generate images. 

%We can easily observe that when $t = 0.5$, the interpolate point is exactly the mean of two embeddings. Also, we note that the geodesic and mean operator on hyperbolic space will reduce the radius of the embeddings due to the change of conformal factor and distance density. Therefore, we further add a re-scale step after interpolation using Möbius scalar multiplication.

\begin{figure}[t]
  \centering
  \setlength{\abovecaptionskip}{0.2cm}   %调整图片标题与图距离
  \setlength{\belowcaptionskip}{0cm}   %调整图片标题与下文距离
   \includegraphics[width=1\linewidth]{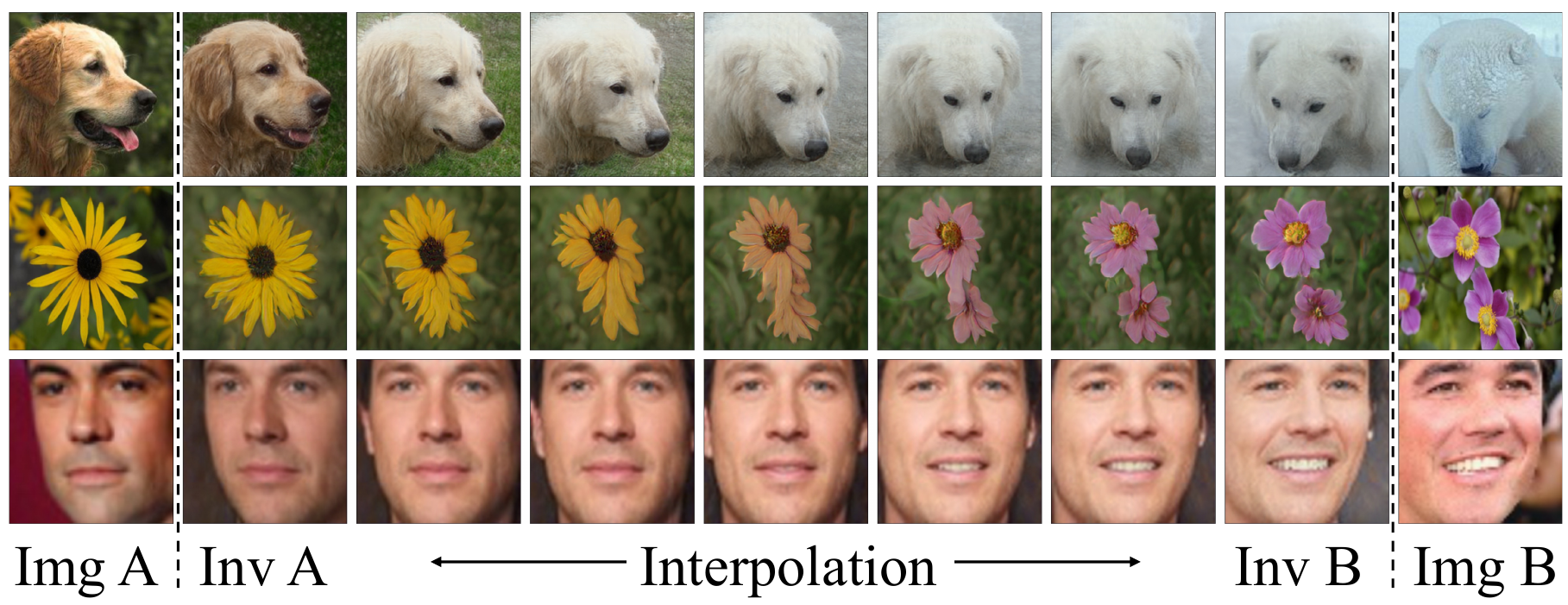}

   \caption{Interpolations in hyperbolic space along the edge of the Poincaré disk (with $r_\mathbb{D} = 6.2126$) on three datasets.}
   \label{fig:interpolation}
\end{figure}

\begin{figure}[t]
  \centering
  \vspace{-0.2cm}  %调整图片与上文的垂直距离
  \setlength{\abovecaptionskip}{0.2cm}   %调整图片标题与图距离
  \setlength{\belowcaptionskip}{-0.2cm}   %调整图片标题与下文距离
   \includegraphics[width=1\linewidth]{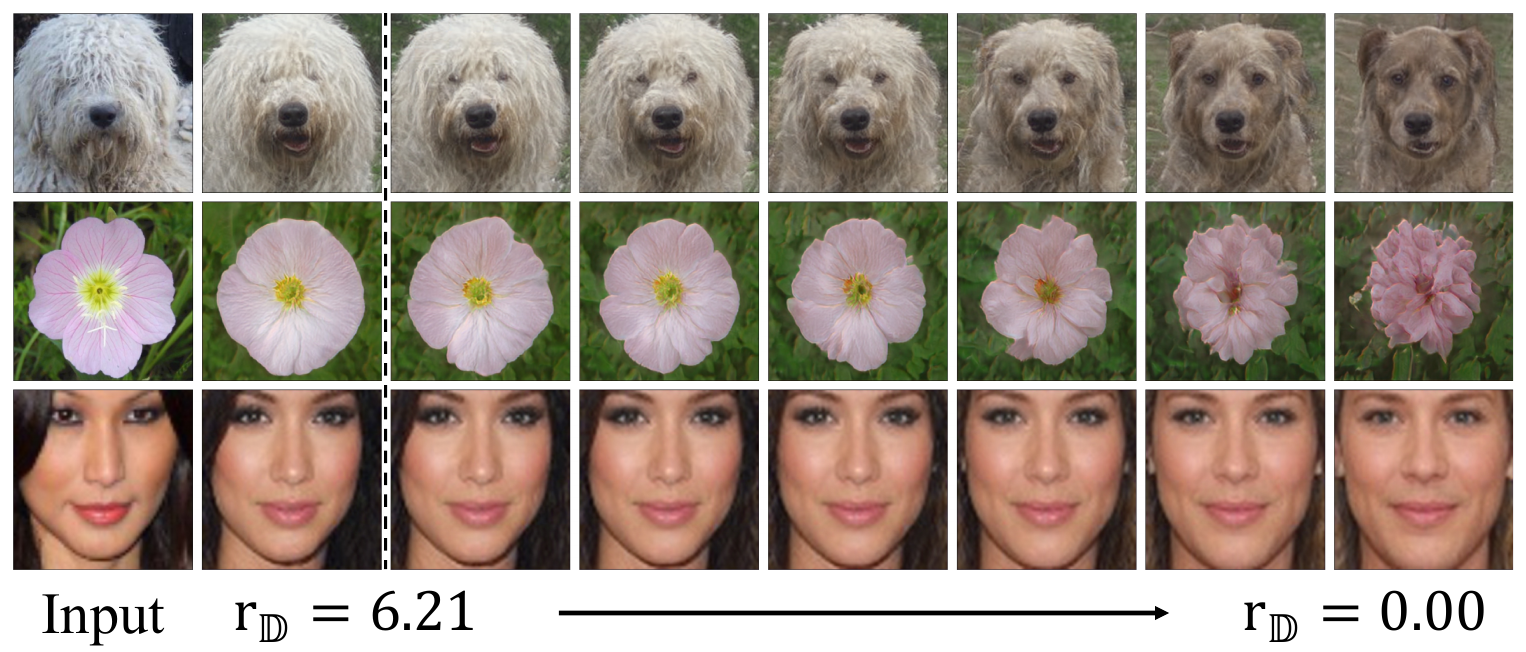}

   \caption{Interpolations by moving the latent codes from the edge to the center of the Poincaré disk (from fine-grained to abstract) on three datasets.}
   \label{fig:interpolation_center}
\end{figure}

\section{Experiment}
\subsection{Implementation Details}
In the training stage, we first train a StyleGAN2~\cite{Karras19} and pSp~\cite{Richardson21} with seen categories. Given a trained pSp, the MLP encoder $\mathtt{MLP}_E$ is an 8-layer MLP with a Leaky-ReLU activation function. The dimension of the latent code in hyperbolic space is chosen to be $512$. More details can be found in the supplementary.

\subsection{Datasets}
We evaluate our method on Animal Faces~\cite{Liu19Few}, Flowers~\cite{Nilsback08}, and VGGFaces~\cite{Parkhi15} following the settings described in~\cite{Ding22}. 

\noindent \textbf{Animal Faces}. We randomly select 119 categories as seen for training and leave 30 as unseen categories for testing.

\noindent \textbf{Flowers}. The Flowers~\cite{Nilsback08} dataset is split into 85 seen categories for training and 17 unseen categories for testing.

\noindent \textbf{VGGFaces}. For VGGFaces~\cite{Parkhi15}, we randomly select 1802 categories for training and 572 for testing.

\begin{table*}
    \centering
    \setlength{\abovecaptionskip}{0.2cm}   %调整图片标题与图距离
    \setlength{\belowcaptionskip}{-0.4cm}   %调整图片标题与下文距离
    \begin{tabular}{p{2.8cm} p{2cm}<{\centering} m{1.5cm}<{\centering} m{1.5cm}<{\centering} m{1.5cm}<{\centering} m{1.5cm}<{\centering} m{1.5cm}<{\centering} m{1.5cm}<{\centering}}
        \toprule
        \multicolumn{1}{l}{\multirow{2}{*}{Method}} & \multicolumn{1}{c}{\multirow{2}{*}{Settings}} & \multicolumn{2}{c}{Flowers} & \multicolumn{2}{c}{Animal Faces} & \multicolumn{2}{c}{VGG Faces*} \\ 
        & & FID($\downarrow$) &  LPIPS($\uparrow$) & FID($\downarrow$) &  LPIPS($\uparrow$) & FID($\downarrow$) &  LPIPS($\uparrow$)\\
        \midrule
        DAWSON~\cite{Liang20} & $3$-shot & 188.96 & 0.0583 &208.68 & 0.0642 & 137.82 & 0.0769\\
        MatchingGAN~\cite{Hong20Match} & $3$-shot & 143.35 & 0.1627 &148.52 & 0.1514 & 118.62 & 0.1695\\
        F2GAN~\cite{Hong20F2} & $3$-shot & 120.48 & 0.2172 &117.74 & 0.1831 & 109.16 & 0.2125\\
        LoFGAN~\cite{Gu21} & $3$-shot & 79.33 & 0.3862 &112.81 & 0.4964 & \textbf{20.31} & 0.2869\\
        \midrule
        DeltaGAN~\cite{Hong22Delta} & $1$-shot & 109.78 & 0.3912 &89.81 & 0.4418 & 80.12 & 0.3146\\
        Disco-FUNIT~\cite{Hong22} & $1$-shot & 90.12 & 0.4436 &71.44 & 0.4511 & - & -\\
        AGE~\cite{Ding22} & $1$-shot & \underline{45.96} & 0.4305 & 28.04 & \underline{0.5575} & \underline{34.86} & \underline{0.3294}\\
        SAGE~\cite{Ding23} & $1$-shot & \textbf{43.52} & \underline{0.4392} &\underline{27.43} & 0.5448 & 34.97 & 0.3232\\
        \midrule
        HAE (Ours) & $1$-shot & 50.10 & \textbf{0.4739} & \textbf{26.33} & \textbf{0.5636} & 35.93 & \textbf{0.5919}\\
        \bottomrule
    \end{tabular}
    \caption{FID($\downarrow$) and LPIPS($\uparrow$) of images generated by different methods for unseen categories on three datasets. \textbf{Bold} indicates the best results and \underline{underline} indicates the second best results. VGGFaces is marked with * because different methods report different numbers of unseen categories on this dataset(\textit{e.g.} 552 in LoFGAN, 96 in DeltaGAN, 497 in L2GAN, and 572 in AGE and SAGE). Note that: Disco-FUNIT~\cite{Hong22} does not provide pre-trained models on VGG Faces~\cite{Parkhi15} dataset.}
    \label{tab:quantitative_comparison}
\end{table*}

\begin{figure*}[t]
  \centering
  \setlength{\abovecaptionskip}{0.2cm}   %调整图片标题与图距离
  \setlength{\belowcaptionskip}{-0.2cm}   %调整图片标题与下文距离
   \includegraphics[width=1\linewidth]{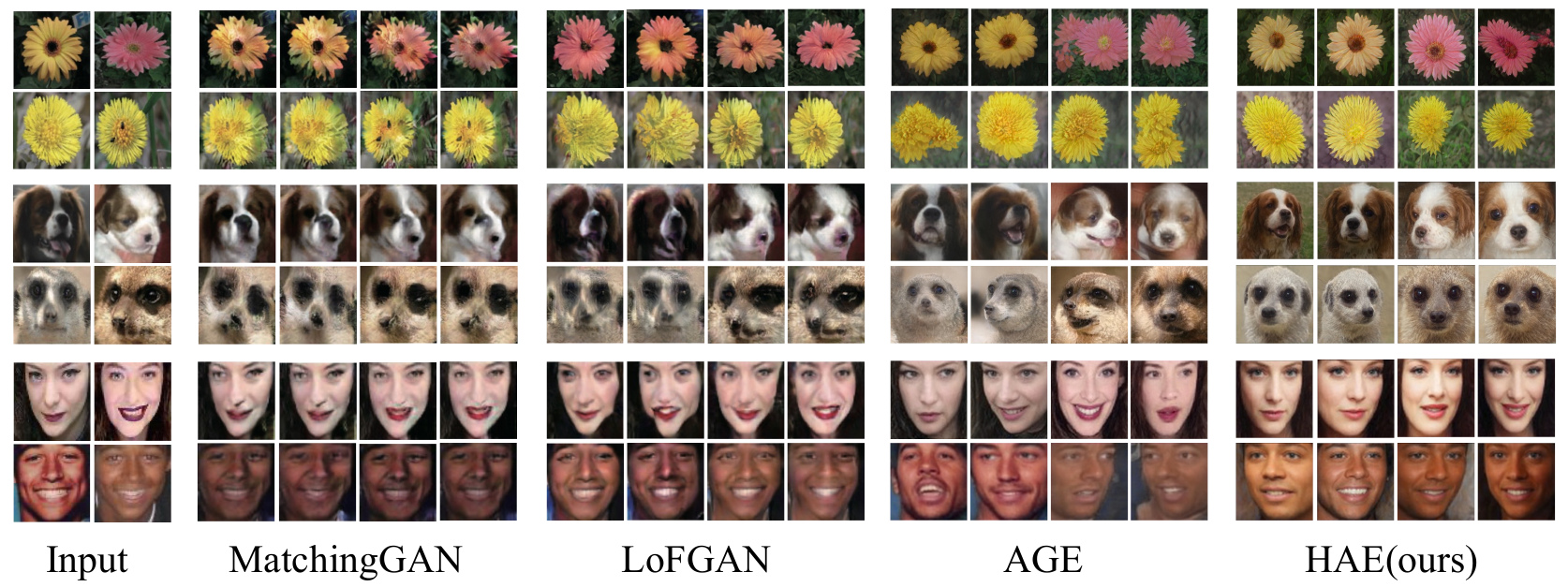}

   \caption{Comparison between images generated by MatchingGAN, LoFGAN, AGE, and HAE on Flowers, Animal Faces, and VGGFaces. Zoom in to see the details. Note that: SAGE~\cite{Ding23} has not released code and pre-trained models. }
   \label{fig:Comparison}
\end{figure*}

\subsection{Analysis of Hierarchical Feature Editing}
\label{sec: Analysis of Hierarchical Feature Editing}
We analyze the properties of the learned hierarchical representations and how the levels of attributes relate to their locations of latent codes in hyperbolic space.

As we mentioned in \cref{sec:Hierarchical Representation}, there is a continuum from fine-grained attributes to abstract attributes, corresponding to the points from the peripheral to the center of the ball. We define the hyperbolic radius $r_\mathbb{D}$\footnote{The radius of the Poincaré disk in our experiment is about $6.2126$} as the hyperbolic distance of the given latent code to the center of the Poincaré disk. To study the influence of the radius of embeddings in hyperbolic space, we run several experiments with different choices of $r_\mathbb{D}$.

\noindent \textbf{Hyperbolic Perturbation and Interpolation.} 
As mentioned in \cref{Image Generation}, we demonstrate the results of perturbation and interpolation.
In addition to the choice of perturbation, we can set the intensity of the perturbation by controlling both the step distance and radius as shown in \cref{fig:perturbation}. The results show that level of semantic attributes is highly related to $r_\mathbb{D}$. With the radius becoming smaller, the attributes become more abstract. We further visualize this property of hyperbolic space by moving the latent codes of the given image from the edge of the Poincaré disk to the center. As \cref{fig:interpolation_center} shows, the images change from very fine-grained to very abstract (the average of all images). The results in \cref{fig:interpolation} show that we can achieve smooth interpolation in hyperbolic space without any distortion. The results demonstrate that with HAE, we can freely control the editing geodesically and hierarchically.

\subsection{Few-shot Image Generation}
\begin{figure}[t]
  \centering
  \setlength{\abovecaptionskip}{0.2cm}   %调整图片标题与图距离
  \setlength{\belowcaptionskip}{-0.5cm}   %调整图片标题与下文距离
   \includegraphics[width=0.9\linewidth]{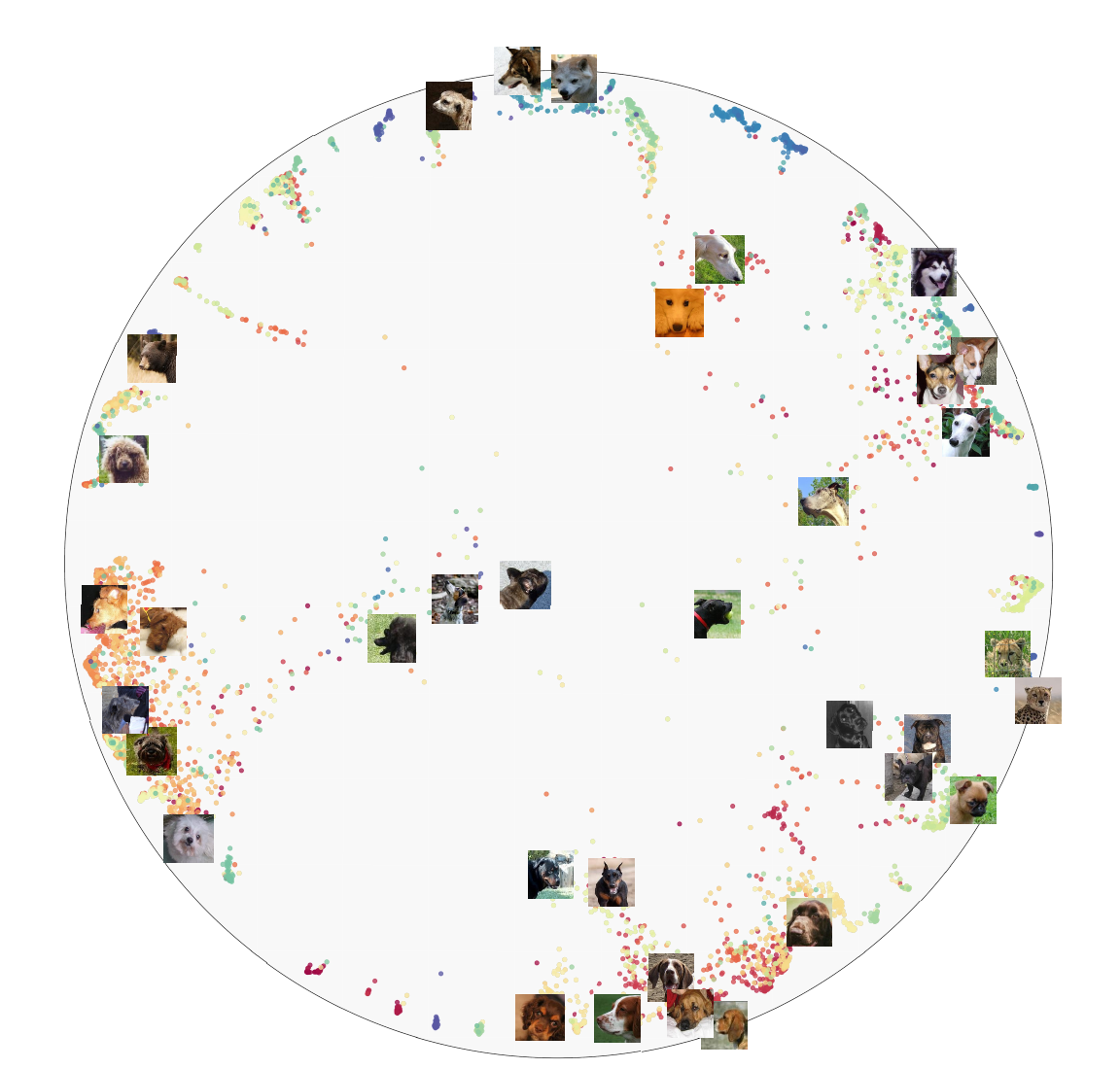}

   \caption{UMAP visualization of hyperbolic 2-D embeddings of Animal Faces dataset. We observe that similar categories are clustered and positioned near the boundary, while ambiguous samples are located near the center. Zoom in to see the details.}
   \label{fig:umap}
\end{figure}

 As \cref{fig:perturbation} shows, the image categories will be changed when $r_\mathbb{D}$ is smaller than about $4$, and the category-irrelevant attributes of images will be changed when $r_\mathbb{D}$ is larger than about 5. The embeddings of Animal Faces are visualized in 2-D Poincaré disk using UMAP~\cite{mcinnes2018umap-software} shown in \cref{fig:umap}. As \cref{fig:interpolation} shows, the posture and the angle of the images will be changed at the early stage of interpolation without changing the category. Thus, the images can be generated by moving the latent code of a given image to some randomly selected semantic direction within the cluster of the category. In practice, we select $r_\mathbb{D}=6.21$ and step size of perturbation as 8 to achieve few-shot image generation as \cref{fig:generation} shows the diverse images generated by adding random perturbations from seen categories. We conduct three experiments to show that HAE can achieve promising few-shot image generation. More examples of generated images are available in the supplementary.

\noindent \textbf{Quantitative Comparison with State-of-the-art.} We calculate the FID~\cite{Heusel17} and LPIPS~\cite{Zhang18unreasonable} to evaluate the fidelity and diversity of the generated images following one-shot settings in~\cite{Ding22, Ding23}. The comparison results are shown in \cref{tab:quantitative_comparison}. Our method achieves the best scores on most of the FID and LPIPS metrics compared with state-of-the-art few-shot image generation methods, which indicates that our method not only improves the model from the semantic aspect but also achieves state-of-the-art performance on the traditional evaluation metrics. Specifically, the LPIPS score of HAE beats SOTA model SAGE on all three datasets since HAE can generate more diverse images.

\noindent \textbf{Qualitative Evaluation.} We qualitatively compare our method with MatchingGAN~\cite{Hong20Match}, LoFGAN~\cite{Gu21}, DeltaGAN~\cite{Hong22Delta} and AGE~\cite{Ding22}. As shown in \cref{fig:Comparison}, HAE can generate images with diversity and fine-grained details. More importantly, the newly generated images have more semantic diversity than others. For instance, the shadow and skin color of the generated faces change with the light condition, and this effect looks more natural. We further conduct a user study by randomly selecting 60 (20 from each dataset) images with generated variants using AGE and HAE. 50 users from different backgrounds are asked to rate the results only based on diversity and quality external information. This is achieved by randomly shuffling the order of images pairwisely and inside any pair. HAE won by a ratio of $58.1\%$ $(1743/3000)$ over AGE (more details in supplementary).

%We also suggest that using a small radius for interpolation can generate meaningful abstract parents for a further step-by-step generation for fine-grained children with desired features.

\begin{figure}[t]
  \centering
  \setlength{\abovecaptionskip}{0.2cm}   %调整图片标题与图距离
  \setlength{\belowcaptionskip}{-0.2cm}   %调整图片标题与下文距离
   \includegraphics[width=0.9\linewidth]{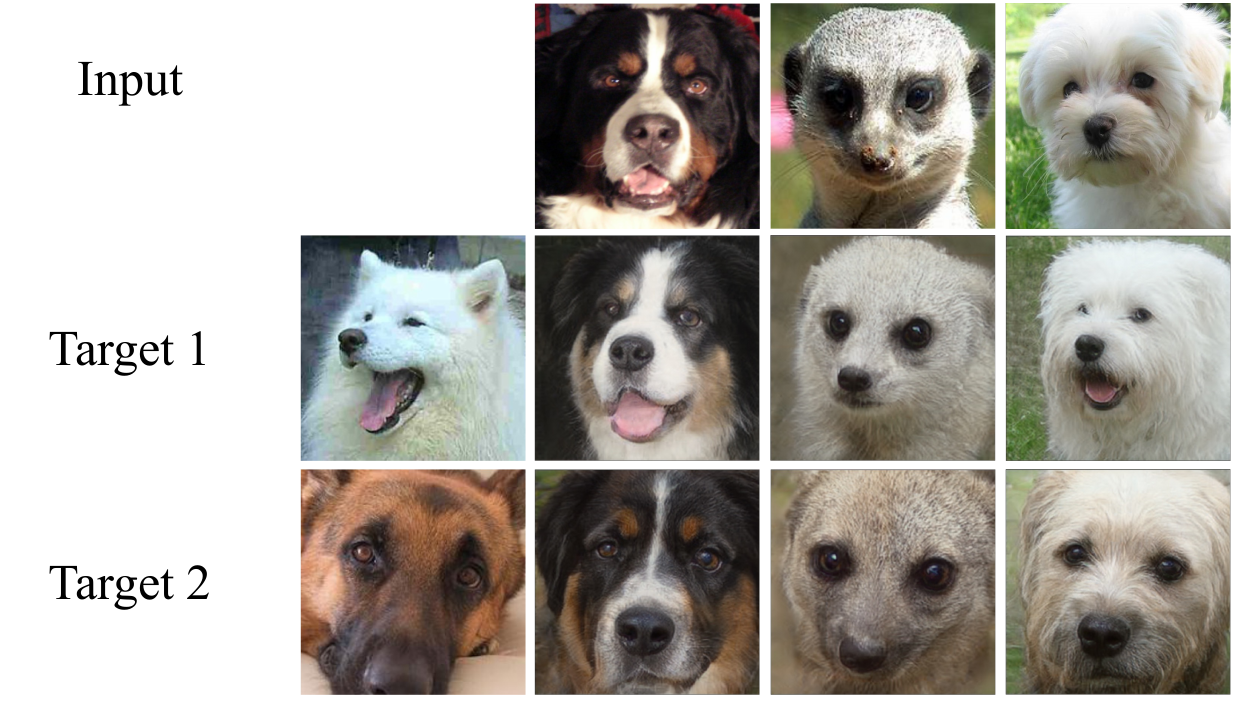}

   \caption{Manipulate images from different categories with the same perturbation (Target 1\&2).}
   \label{fig:Transferability}
\end{figure}

\noindent \textbf{Transferability.} If we move latent codes at category-irrelevant levels, the target perturbation is transferable across all categories. We edit the images from three categories with the same editing direction, the output images are shown in \cref{fig:Transferability}. It demonstrates that HAE achieves a highly controllable and interpretable editing process.

%1. few-shot image classification /done
%2. visualize Poincare ball /done
%3. few-shot image generation examples /done
%4. Hyperbolic perturbation visualization /ignore
%5. Inference of different level of latent codes /ignore
%6. Manipulate images from different categories with the same /ignore perturbation /done

%1.降维可操作性高，性能几乎不损失。
%2.hierarchical 

\begin{table}[h]\footnotesize
    \centering
    \setlength{\abovecaptionskip}{0.2cm}   %调整图片标题与图距离
    \setlength{\belowcaptionskip}{-0.2cm}   %调整图片标题与下文距离
    \begin{tabular}{p{1cm}m{0.7cm}<{\centering} m{0.7cm}<{\centering} m{0.7cm}<{\centering} m{0.7cm}<{\centering} m{0.7cm}<{\centering} m{0.7cm}<{\centering}}
        \toprule
        \multicolumn{1}{l}{\multirow{2}{*}{Method}} & \multicolumn{2}{c}{Flowers} & \multicolumn{2}{c}{Animal Faces} & \multicolumn{2}{c}{VGG Faces} \\ 
        & FID &  LPIPS & FID &  LPIPS & FID &  LPIPS\\
        \midrule
        SAGE~\cite{Ding23} & \textbf{43.52} & \underline{0.4392} &27.43 & \underline{0.5448} & \textbf{34.97} & 0.3232\\
        \midrule
        HAE(Euc) & 54.62 & 0.4293 & \textbf{25.27} & 0.5129 & 38.46 & \underline{0.5908}\\
        HAE(Hyp) & \underline{50.10} & \textbf{0.4739} & \underline{26.33} & \textbf{0.5636} & \underline{35.93} & \textbf{0.5919}\\
        \bottomrule
    \end{tabular}
    \caption{FID($\downarrow$) and LPIPS($\uparrow$) of images generated by HAE in different geometries for unseen categories on three datasets. \textbf{Bold} indicates the best results and \underline{underline} indicates the second best results.}
    \label{tab:ablation_euc_hyp}
\end{table}

\subsection{Ablation Study}

\noindent \textbf{HAE in Euclidean.} We re-trained HAE models in Euclidean space with the NLL loss to validate the performance gain in \cref{tab:quantitative_comparison} is due to the hierarchical hyperbolic representation rather than the disentanglement caused by \cref{eq:NLLLoss}. The quantitative comparison is shown in \cref{tab:ablation_euc_hyp}. It shows that the hyperbolic space boosts the performance, especially for the LPIPS score, since the latent code is more disentangled in hyperbolic space~\cite{Ge22}. This finding is also supported by the UMAP visualization in ~\cref{fig:umap}. More details can be found in the supplementary material.

\begin{figure}[t]
  \centering
  \setlength{\abovecaptionskip}{0.1cm}   %调整图片标题与图距离
  \setlength{\belowcaptionskip}{-0.2cm}   %调整图片标题与下文距离
   \includegraphics[width=1\linewidth]{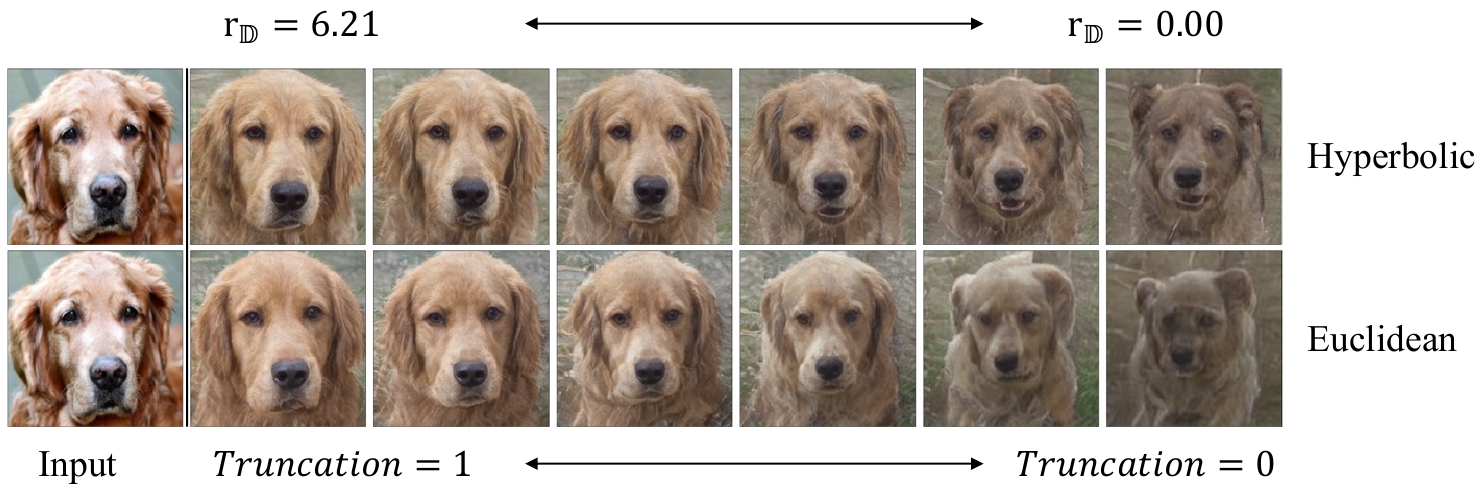}
   \caption{\textbf{Top}: Interpolations by moving the latent codes from the edge to the center in hyperbolic space. \textbf{Bottom}: Interpolation with different truncation in Euclidean space. Zoom in to see the details.}
   \label{fig:Hyp_Euc_radius}
\end{figure}

\begin{figure}[t]
  \centering
  \setlength{\abovecaptionskip}{0.1cm}   %调整图片标题与图距离
  \setlength{\belowcaptionskip}{-0.2cm}   %调整图片标题与下文距离
   \includegraphics[width=1\linewidth]{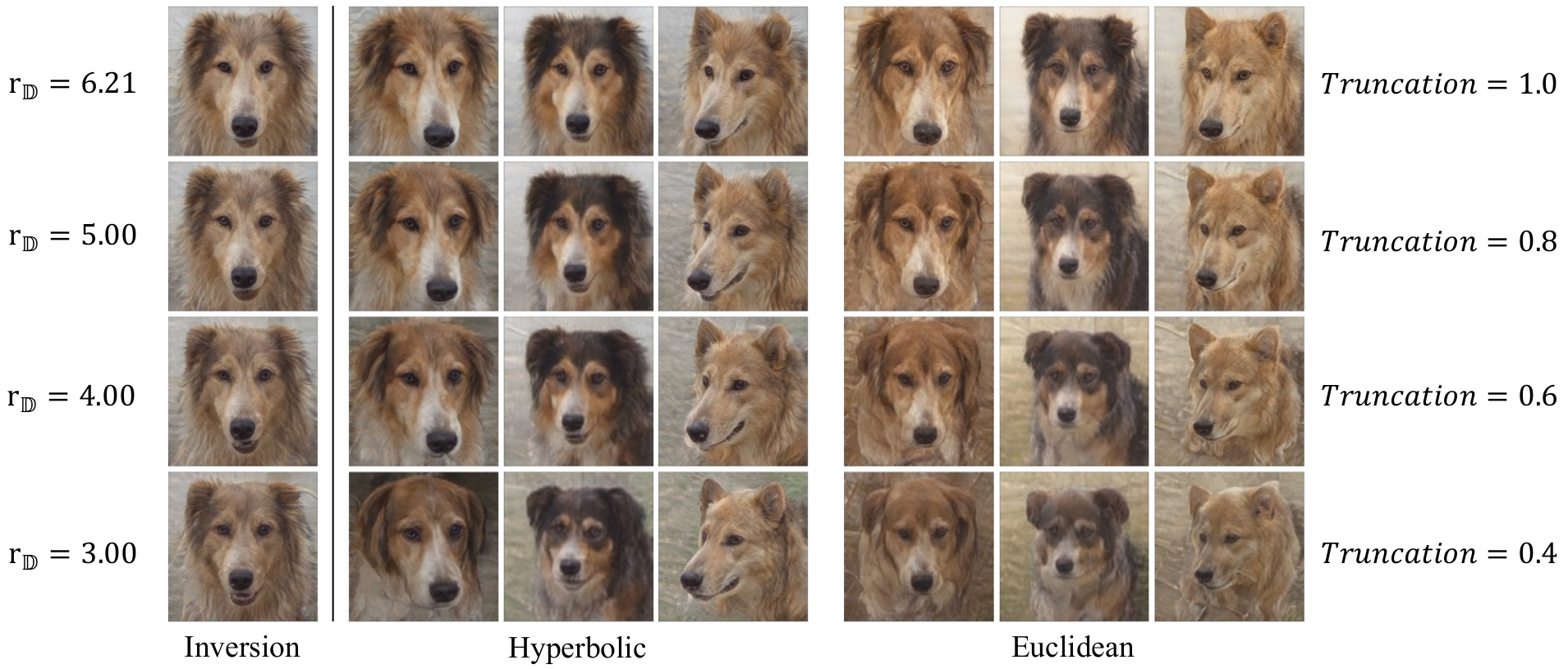}

   \caption{Images generated by HAE(Hyp) and HAE(Euc) by adding the same perturbation on the latent code of a given image with different settings of hyperbolic radius and truncation. Zoom in to see the details.} 
   \label{fig:Hyp_Euc_perturbation}
\end{figure}

\noindent \textbf{Hyperbolic Radius versus Truncation.} 
StyleGAN~\cite{Karras19} uses truncation trick~\cite{Marchesi17,Brock19,Karras19,Karras20} in $\mathcal{W}$-space to achieve the balance between the image quality and diversity. The experiments in ~\cite{Karras19, Karras20} also show that the truncation level in $\mathcal{W^+}$-space control the level of abstraction of the generated images. We conduct the experiments in~\cref{Image Generation} using truncation to validate the gains of hyperbolic space. The results are illustrated in \cref{fig:Hyp_Euc_radius} and \cref{fig:Hyp_Euc_perturbation}. As \cref{fig:Hyp_Euc_radius} shows, the category of the image changes along with the posture of the dog as the truncation gets smaller, while the category-relevant attributes do not change when the hyperbolic radius ($r_\mathbb{D}$) is large. This can also be proved in \cref{fig:Hyp_Euc_perturbation}. The category remains the same after adding perturbation when $r_\mathbb{D}$ is large, while the truncation can not control semantic-level editing. This shows that Euclidean space can only capture scale-based hierarchy rather than the semantic hierarchy.

\noindent \textbf{Downstream Task.} 
We conduct data augmentation via HAE for image classification on Animal Faces~\cite{Parkhi15}. We randomly select 30, 35, and 35 images for each category as train, val, and test, respectively. Following~\cite{Gu21}, a ResNet-18 backbone is initialized from the seen categories. Then the model is fine-tuned on the unseen categories referred to as the \textit{baseline}. 60 images are generated for each unseen category as data augmentation.
The result is presented in \cref{tab:ablation_downstream_task}. The diversity and quality of generated images are primarily controlled by the hyperbolic radii $r_\mathbb{D}$. As the radius becomes smaller, HAE generates images of higher diversity, but categories (referring to high-level attributes) also gradually change to others. $r_\mathbb{D}=6$ achieves the best performance on the classification experiment. However, the performance drops when the radius is smaller than 4.5. This is because the semantic attributes change too much and thus mislead the classifier.

\subsection{Limitations and Future Work}
Although HAE achieves reliable hierarchical attribute editing in hyperbolic space for few-shot image generation, there are several limitations.

First, the boundary of category changing is hard to be quantified since the hierarchical levels are continuous in the hyperbolic space. Users need to find a ``safe" boundary by trying different radii and step sizes of perturbation before generating new images. However, from another perspective, this continuity of hierarchy provides flexibility for users to set different boundaries for different downstream tasks as they need.
    
Second, the performance of HAE is limited by the pretrained styleGAN and the inversion method. If the input image can not be well embedded, the editing will also fail. This problem can be solved by changing more powerful backbones,~\textit{e.g.}, ViT~\cite{Dosovitskiy21}, in future work.
    
Finally, we use supervised learning to get the hierarchical embedding in hyperbolic space. However, the number of images in existing datasets with labels for generation tasks is relatively small, which makes the embeddings in the hyperbolic space not evenly distributed. The solution to this problem is simple, use unsupervised learning with large-scale high-quality datasets.

\begin{table}
    \centering
    \setlength{\abovecaptionskip}{0.2cm}   %调整图片标题与图距离
    \setlength{\belowcaptionskip}{-0.2cm}   %调整图片标题与下文距离
    \begin{tabular}{p{1.65cm}<{\centering} m{1.65cm}<{\centering} m{1.65cm}<{\centering} m{1.65cm}<{\centering}}
        \toprule
        Hyperbolic Radius & {\multirow{2}{*}{Accuracy}} & {\multirow{2}{*}{FID($\downarrow$)}} & {\multirow{2}{*}{LPIPS($\uparrow$)}} \\ 
        \midrule
        baseline & 58.67 & - & - \\
        \midrule
        6.0 & \textbf{60.10} & \textbf{46.89} & 0.4520 \\
        5.5 & 59.52 & 48.68 & 0.4651 \\
        5.0 & 59.05 & 52.08 & 0.4823 \\
        4.5 & 59.14 & 60.87 & 0.5174 \\
        4.0 & 58.57 & 65.83 & 0.5386 \\
        3.5 & 56.86 & 68.44 & 0.6034 \\
        3.0 & 54.14 & 69.40 & \textbf{0.6316} \\
        \bottomrule
    \end{tabular}
    \caption{Ablation of same perturbation on different radii on Animal Faces.}
    \label{tab:ablation_downstream_task}
\end{table}

\section{Conclusion}
In this work, we propose a simple yet effective method HAE to edit hierarchical attributes in hyperbolic space. After learning the semantic hierarchy from images, our model is able to edit continuous semantic hierarchical features of images for flexible few-shot image generation in the hyperbolic space. Experiments demonstrate that HAE is capable of  achieving not only stable few-shot image generation with state-of-the-art quality and diversity but a controllable and interpretable editing process. Future work includes the combination of HAE and large pretrained models and applications to more downstream tasks.
\\
\\
\textbf{Acknowledgement.} This work was supported in part by the National Key R\&D Program of China under Grant 2018AAA0102000, in part by National Natural Science Foundation of China: 62022083 and 62236008.

%appendix content
%1. visualization of UMAP on W+ space, hyperbolic embedding of other
%   two datasets
%2. comparison of generated images with AGE
%3. More generation examples
%4. size of parameters and computing resources required analysis
%5. Ablation study of downstream tasks on other two datasets
%6. Mathematical formulas
%7. More implementation details
%8. More interpolation visualization

%%%%%%%%% REFERENCES
{\small
\bibliographystyle{ieee_fullname}
\bibliography{egbib}
}

\clearpage
\begin{appendix}
\section*{ \huge Supplementary Material\vspace{1em}}

\section*{Overview}
This appendix is organized as follows: 

\cref{Hyperbolic Neural Networks} provides the mathematical formulae used in hyperbolic neural networks. Sec 3.2 \& Sec 3.3

\cref{Implementation Details and Analysis} gives more implementation details of HAE. Sec 4.1

\cref{Ablation Study of Downstream Task} shows the results of the ablation study of downstream tasks for Animal Faces~\cite{Liu19Few}, Flowers~\cite{Nilsback08} and VGGFaces~\cite{Parkhi15}. Sec 4.3

\cref{Comparison with Euclidean space} compares the embeddings of images in hyperbolic space and Euclidean space. Sec 4.4

\cref{Interpolation Visualization} visualizes the interpolation on different radii in the Poincaré disk, along the geodesic, and on $\mathcal{W^+}$-space. Sec 4.3

\cref{Images Generated with Different Radii} shows the images generated with different radii in the Poincaré disk. Sec 4.3

\cref{Comparison with State-of-the-art few-shot image generation Method} compares the images generated by state-of-the-art few-shot image generation method, \ie AGE~\cite{Ding22} and our methods HAE. Sec 4.4

\cref{User Study} gives more details of the user study we conducted. Sec 4.4

\cref{Additional Examples Generated by HAE} gives more examples generated by HAE. Sec 4.4

\section{Hyperbolic Neural Networks}
\label{Hyperbolic Neural Networks}
For hyperbolic spaces, since the metric is different from Euclidean space, the corresponding calculation operators also differ from Euclidean space. Recall that in Eq. (11), we have two operations: Möbius addition and Möbius scalar multiplication~\cite{Khrulkov21}, given fixed curvature $c$.

For any given vectors $x, y \in \mathbb{H}^n$, the \textit{Möbius addition} is defined by:

\begin{equation}
\label{mobius addition}
x \oplus_c y=\frac{\left(1-2 c\langle x, y\rangle-c\|y\|_2^2\right) x+\left(1+c\|x\|_2^2\right) y}{1-2 c\langle x, y\rangle+c^2\|x\|_2^2\|y\|_2^2},
\end{equation}
where $\| \cdot \| $ denotes the $2$-norm of the vector, and $\langle \cdot, \cdot \rangle$ denotes the Euclidean inner product of the vectors.

Similarly, the \textit{Möbius scalar multiplication} of a scalar $r$ and a given vector $x \in \mathbb{H}^n$ is defined by:
\begin{equation}
r \otimes_c x=\tan _c\left(r \tan _c^{-1}\left(\|x\|_2\right)\right) \frac{x}{\|x\|_2}.
\end{equation}

We also would like to give explicit forms of the exponential map and the logarithmic map which are used in our model to achieve the translation between hyperbolic space and Euclidean space as mentioned in Sec 3.2.

The \textit{exponential map} $\exp _{x}^c: T_{x} \mathbb{D}_c^n \cong \mathbb{R}^n \to \mathbb{D}_c^n$, that maps from the tangent spaces into the manifold, is given by
\begin{equation}
    \exp _{x}^c(v):=x \oplus_c\left(\tanh \left(\sqrt{c} \frac{\lambda_{x}^c\|v\|}{2}\right) \frac{v}{\sqrt{c}\|v\|}\right).
\end{equation}

The \textit{logarithmic map} $\log _{x}^c(y): \mathbb{D}_c^n \to T_{x} \mathbb{D}_c^n \cong \mathbb{R}^n$ is given by
\begin{equation}
    \log _{x}^c(y):=\frac{2}{\sqrt{c} \lambda_{x}^c} \operatorname{arctanh}\left(\sqrt{c}\left\|-x \oplus_c y\right\|\right) \frac{-x \oplus_c y}{\left\|-x \oplus_c y\right\|}.
\end{equation}

\section{Implementation Details and Analysis}
\label{Implementation Details and Analysis}
As mentioned in Sec 3.2, the output of $\mathtt{pSp}$: $\mathbf{w}_i \in \mathbb{R}^{18 \times 512}$. The MLP encoder $\mathtt{MLP}_E$ used in HAE, is split into three parts: $\mathrm{encoder}_{low}$, $\mathrm{encoder}_{mid}$, and $\mathrm{encoder}_{high}$ for encoding lower layer attributes, middle layer attributes, and higher layer attributes. The $\mathrm{encoder}_{low}$ is a 5-layer MLP with a Leaky-ReLU (slope=0.2) activation function. The first three layers of $\mathbf{w}_i$ are then fed into $\mathrm{encoder}_{low}$. The dimension of the output attribute is 128. Similar to $\mathrm{encoder}_{low}$, the $\mathrm{encoder}_{mid}$ is also an 5-layer MLP. 3-7 layers of $\mathbf{w}_i$ is then fed into $\mathrm{encoder}_{mid}$, the dimension of the output attribute is also 128. Different from $\mathrm{encoder}_{low}$ and $\mathrm{encoder}_{mid}$, $\mathrm{encoder}_{high}$ is an 8-layer MLP. And we fed the last 12 layers attributes of $\mathbf{w}_i$ into it. The dimension of the output attribute of $\mathrm{encoder}_{high}$ is 256. Therefore, the final dimension of the Euclidean latent code $z_{\mathbb{R}i}$ is $128+128+256=512$. While the MLP decoder  $\mathtt{MLP}_D$ is the reversed version of $\mathtt{MLP}_E$, taking $z_{\mathbb{R}i}' \in \mathbb{R}^{512}$ as input and output $\mathbf{w}_i' \in \mathbb{R}^{18 \times 512}$.

During the training process, the constants defined in Eq. (10) are set as  $\lambda_1=1$, $\lambda_3=0.3$, and $\lambda_2$ changes dynamically based on the value of $\mathcal{L}_{\text{rec}}$ which guarantees that $0.6 \geq \lambda_2 \mathcal{L}_{\text{rec}}\geq 0.3$. Besides, we employ Adam optimizer with a learning rate of $1\mathrm{e}{-4}$, and the batch size is set to 8.

In addition, as a remark, we choose the largest radius as $6$ in most of our experiments as in hyperbolic space since any vector asymptotically lying on the surface unit $N$-sphere will have a hyperbolic length of approximately $r = 6.2126$, which can be directly calculated by Eq. (2).
 
Finally, we want to show that HAE can be easily trained. The size of trainable parameters of HAE is around one hundred million which is small compared with other models with billions of parameters. It can be trained well within one day using a single NVIDIA TITAN RTX.

\section{Ablation Study of Downstream Task}
\label{Ablation Study of Downstream Task}
Similar to the ablation study in Sec 4.3. We also conduct data augmentation via HAE for image classification on Animal Faces~\cite{Liu19Few}, Flowers~\cite{Nilsback08}, and VGGFaces~\cite{Parkhi15}. Due to the limited size of Flowers and VGGFaces datasets. We randomly select 10, 15 and 15 images for each category as train, val and test, respectively. Following~\cite{Gu21, Ding22}, a ResNet-18 backbone is initialized from the seen categories, then the model is fine-tuned on the unseen categories referred to as the \textit{baseline}. 30 images are generated for each unseen category as data augmentation.

The result of Animal Faces is shown in \cref{tab-label-animals}. It shows that the accuracy of the classifier improves after using the AdamW optimizer. The experiment result essentially confirms the original result in Sec 4.3. The data augmentation improves the performance of the classifier when the hyperbolic radius $r_\mathbb{D}$ is larger than 4. $r_\mathbb{D}=5$ achieves the best performance on the classification experiment mainly because it achieves the best trade-off between the quality and diversity. However, the performance drops when the radius is smaller than 4. This is because the semantic attributes change too much and thus mislead the classifier.

The result of Flowers is presented in \cref{tab-label-flowers}. Similar to the result of Animal Faces, the diversity and quality of generated images are largely controlled by the hyperbolic radii $r_\mathbb{D}$. As the radius becomes smaller, HAE generates images of higher diversity but categories also gradually change to others. $r_\mathbb{D}=6$ achieves the best performance on the classification experiment.

However, \cref{tab-label-faces} shows that all accuracy drops when we do data augmentation on VGGFaces dataset. We estimate that this is due to the low quality of inversion that harms the performance of the classifier. Besides, since we only select 10 images for each category for training, with the limited size of VGGFaces, it is easy to overfit. To evaluate our estimation, we further test the performance of the classifier trained by the original images and inversion images without any perturbation, denoted as \textit{inversion} in \cref{tab-label-faces}. This result proves that our estimation is correct. The accuracy of the classifier trained by augmented images increases compared with the \textit{inversion}, which shows that the augmentation still works and improves the generalization performance of our classifier. $r_\mathbb{D}=3$ achieves the best performance on the classification experiment except the \textit{baseline}.

It is also worth noticing that, the FID and LPIPS scores in\cref{tab-label-animals}, \cref{tab-label-flowers}, and \cref{tab-label-faces} are different from the scores we calculated in Sec 4.4. That is because we only use a very small subset of the data in this ablation study which can not represent the distribution of all images in the test dataset. Besides, the improvement of a classifier trained on augmented data is trivial, we believe this is due to the limitation of the encoding method,\ie psp~\cite{Richardson21} and generator,\ie StyleGAN2~\cite{Karras20}, which can not generate images with high enough quality.
\begin{table}
    \centering
    \begin{tabular}{p{1.65cm}<{\centering} m{1.65cm}<{\centering} m{1.65cm}<{\centering} m{1.65cm}<{\centering}}
        \toprule
        Hyperbolic Radius & {\multirow{2}{*}{Accuracy}} & {\multirow{2}{*}{FID($\downarrow$)}} & {\multirow{2}{*}{LPIPS($\uparrow$)}} \\ 
        \midrule
        baseline & 67.34 & - & - \\
        \midrule
        6.0 & 68.22 & \textbf{46.89} & 0.4520 \\
        5.5 & 68.56 & 48.68 & 0.4651 \\
        5.0 & \textbf{69.33} & 52.08 & 0.4823 \\
        4.5 & 68.22 & 60.87 & 0.5174 \\
        4.0 & 67.67 & 65.83 & 0.5386 \\
        3.5 & 67.33 & 68.44 & 0.6034 \\
        3.0 & 66.89 & 69.40 & \textbf{0.6316} \\
        \bottomrule
    \end{tabular}
    \caption{Ablation of same perturbation on different radii on Animal Faces.}
    \label{tab-label-animals}
\end{table}

\begin{table}
    \centering
    \begin{tabular}{p{1.65cm}<{\centering} m{1.65cm}<{\centering} m{1.65cm}<{\centering} m{1.65cm}<{\centering}}
        \toprule
        Hyperbolic Radius & {\multirow{2}{*}{Accuracy}} & {\multirow{2}{*}{FID($\downarrow$)}} & {\multirow{2}{*}{LPIPS($\uparrow$)}} \\ 
        \midrule
        baseline & 71.76 & - & - \\
        \midrule
        6.0 & \textbf{79.21} & \textbf{94.35} & 0.4640 \\
        5.5 & 77.25 & 98.09 & 0.4871 \\
        5.0 & 75.29 & 97.81 & 0.5110 \\
        4.5 & 78.82 & 97.53 & 0.5330 \\
        4.0 & 75.29 & 97.58 & 0.5499 \\
        3.5 & 73.33 & 101.52 & 0.6152 \\
        3.0 & 72.55 & 105.05 & \textbf{0.6439} \\
        \bottomrule
    \end{tabular}
    \caption{Ablation of same perturbation on different radii on Flowers.}
    \label{tab-label-flowers}
\end{table}

\begin{table}
    \centering
    \begin{tabular}{p{1.65cm}<{\centering} m{1.65cm}<{\centering} m{1.65cm}<{\centering} m{1.65cm}<{\centering}}
        \toprule
        Hyperbolic Radius & {\multirow{2}{*}{Accuracy}} & {\multirow{2}{*}{FID($\downarrow$)}} & {\multirow{2}{*}{LPIPS($\uparrow$)}} \\ 
        \midrule
        baseline & \textbf{77.99} & - & - \\
        \midrule
        inversion & 69.53 & \textbf{25.46} & 0.2325 \\
        \midrule
        6.0 & 71.98 & 26.19 & 0.2702 \\
        5.5 & 72.53 & 26.46 & 0.2887 \\
        5.0 & 72.45 & 26.83 & 0.3080 \\
        4.5 & 72.32 & 26.92 & 0.3258 \\
        4.0 & 72.96 & 27.02 & 0.3405 \\
        3.5 & 74.05 & 26.35 & 0.4044 \\
        3.0 & \underline{74.44} & 25.90 & \textbf{0.4411} \\
        \bottomrule
    \end{tabular}
    \caption{Ablation of same perturbation on different radii on VGGFaces.}
    \label{tab-label-faces}
\end{table}

\section{Comparison with Euclidean space}
\label{Comparison with Euclidean space}
\begin{figure}[!htb]
  \centering
   \includegraphics[width=1\linewidth]{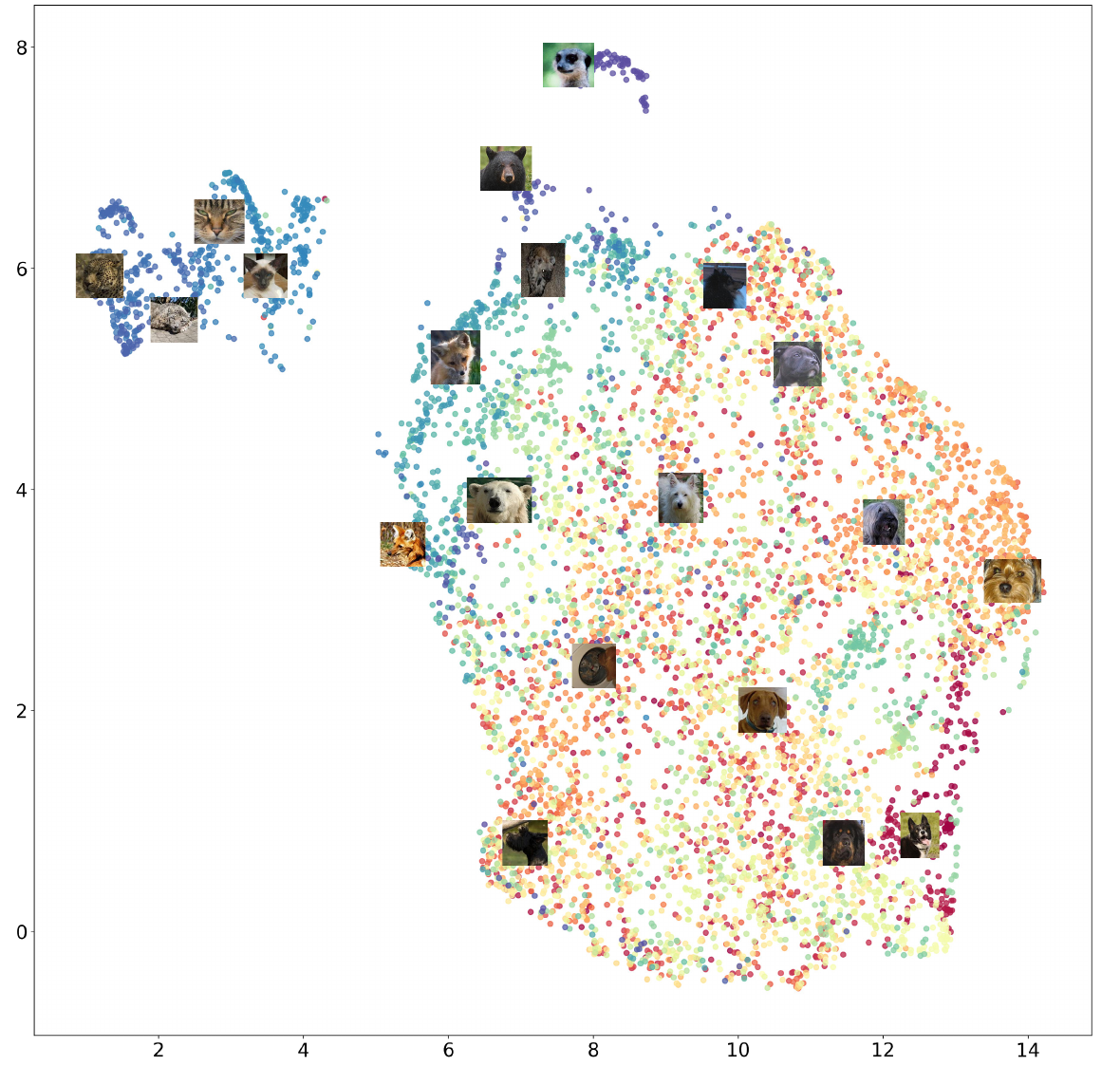}

   \caption{UMAP visualization for Animal Faces dataset, while the embeddings are in the $\mathcal{W^+}$-space of the same model.}
   \label{fig:umap_w_animal_appendix}
\end{figure}

In this section, we mainly compare hyperbolic space with Euclidean space by UMAP visualization~\cite{mcinnes2018umap-software}, which is an extension of our analysis in Sec 4.4. Following the UMAP visualization on hyperbolic space for the Animal Faces~\cite{Parkhi15} dataset, we first show the UMAP visualization of the embeddings of images in $\mathcal{W}^+$-space, where each embedding is of $18\times512$-dimension and therefore we resize them for UMAP calculation. The results for Euclidean UMAP of Animal Faces dataset are shown in \cref{fig:umap_w_animal_appendix}. We observe that although the transition across different categories is smooth, there are no obvious clusters in the plot even for some significantly different species (\eg, polar bears and foxes). 

\begin{figure*}[htb!]
  \centering
   \includegraphics[width=1\linewidth]{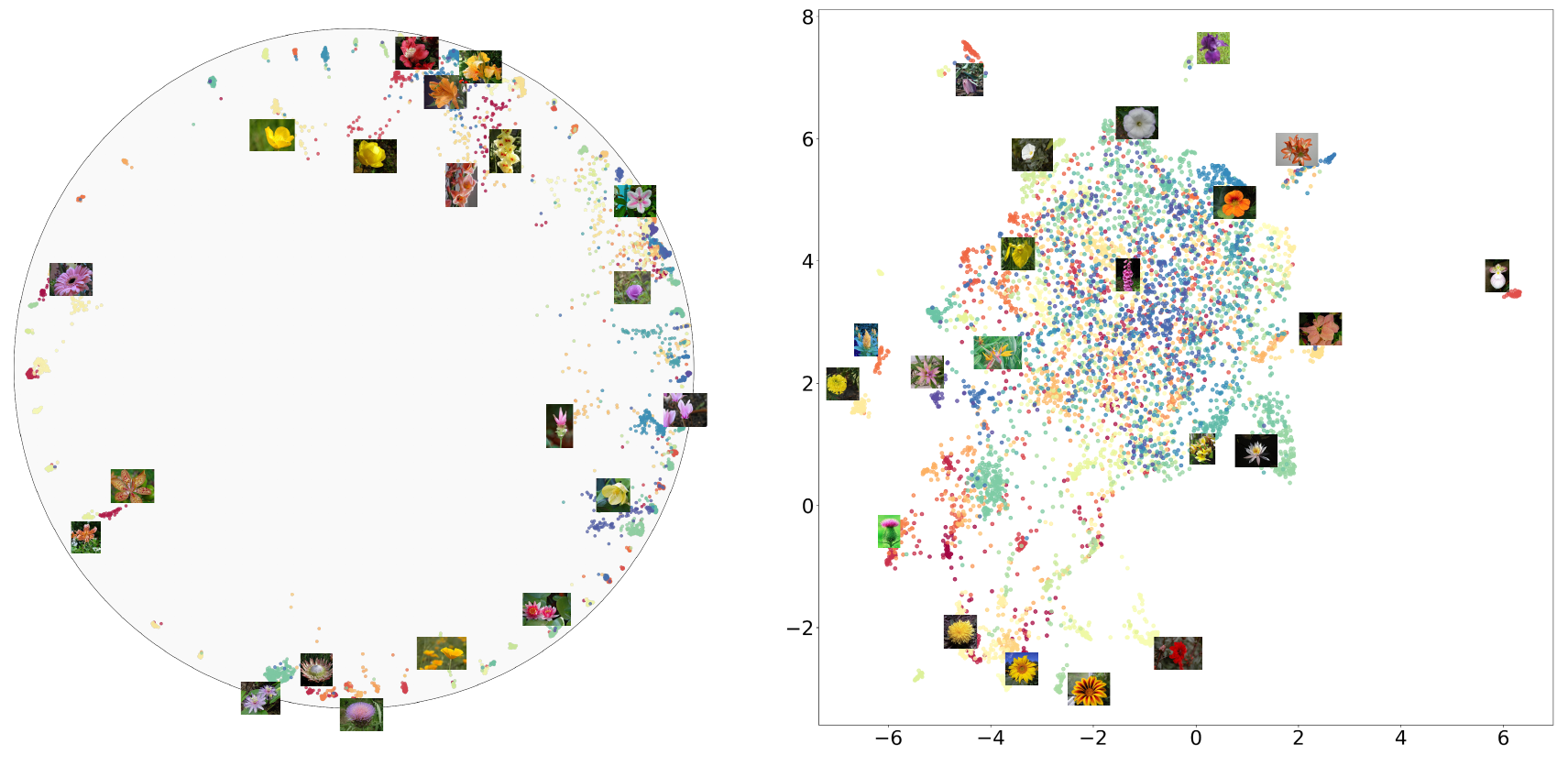}

   \caption{UMAP Visualization for Flowers dataset. Left: Hyperbolic space. Right: Euclidean space ($\mathcal{W}^+$-space).}
   \label{fig:umap_flowers}
\end{figure*}

We further carry on the experiments on the other two datasets. For Flowers~\cite{Nilsback08} dataset, we use all images in the test dataset to generate the embeddings for both spaces, where there are 101 classes in total. The number of images in each class varies due to the original setting of the dataset. The results of the Flowers dataset are shown in \cref{fig:umap_flowers}, where clusters are more obvious in hyperbolic space and the similarity between classes is also well-reflected.

\begin{figure*}[htb!]
  \centering
   \includegraphics[width=1\linewidth]{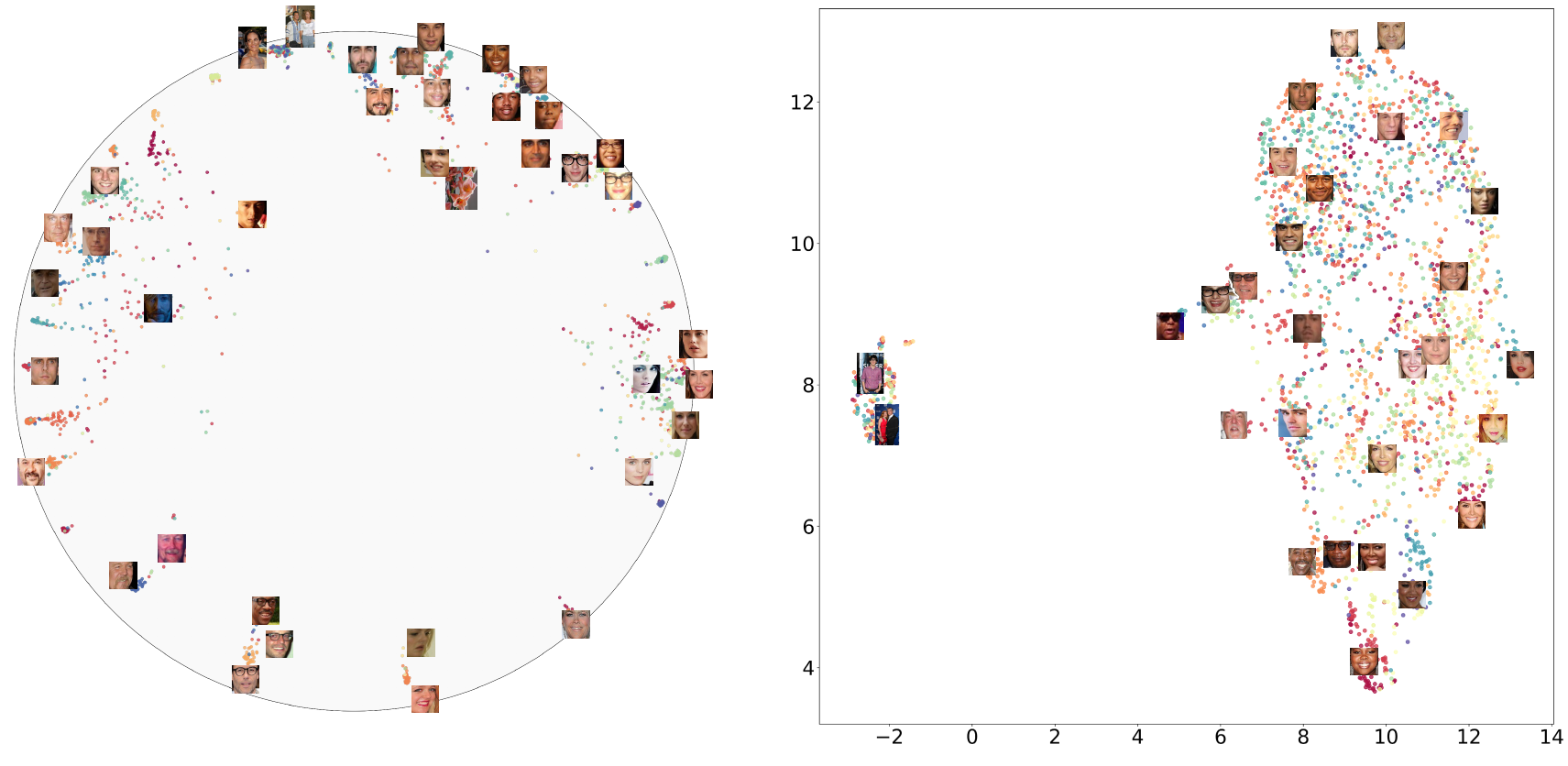}

   \caption{UMAP Visualization for VGGFaces dataset. Left: Hyperbolic space. Right: Euclidean space ($\mathcal{W}^+$-space). In each visualization, no images are from the same category.}
   \label{fig:umap_faces}
\end{figure*}

For the VGGFaces dataset~\cite{Parkhi15}, the results are shown in \cref{fig:umap_faces}. Similarly, the clusters are better represented in hyperbolic settings. We observe that in hyperbolic space, images with similar low-level attributes (\eg, wearing black frame glasses, having mutton chops beard) are clustered. We need to pay attention to the small clusters in both plots where images are represented with tall rectangles (in the plot). These images do not share semantic attributes but are clustered together, which can be the influence of heavy watermarks on the images. This also encourages us to train the model with high-quality datasets for better GAN inversions.

\section{Interpolation Visualization}
\label{Interpolation Visualization}
\begin{figure*}[!htb]
  \centering
   \includegraphics[width=0.75\linewidth]{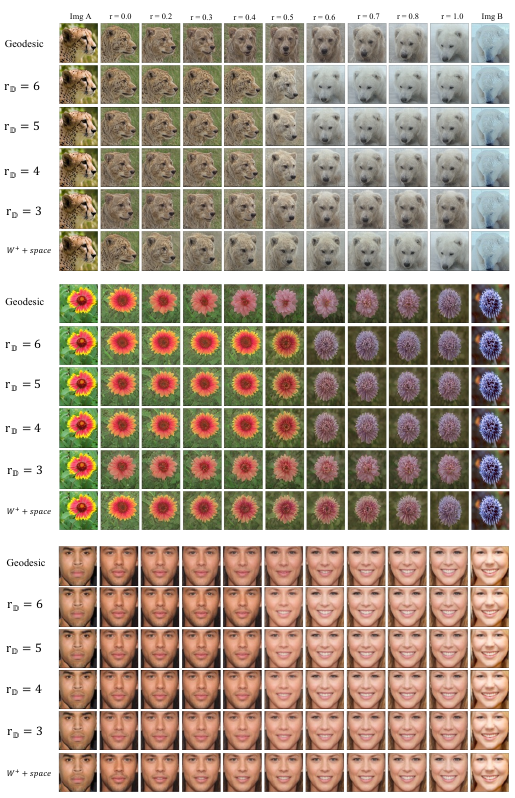}

   \caption{Interpolation along the geodesic and different radii in hyperbolic space and along the straight line in $\mathcal{W}^+$-space on Animal Faces, Flowers, and VGGFaces.}
   \label{fig:interpolation_radius}
\end{figure*}

In this section, we offer a more detailed comparison of different radii and latent spaces. The results are shown in \cref{fig:interpolation_radius}, where $r$ refers to the ratio of the whole geodesic described in Eq. (11) starting from image A to image B, \eg, when $r = 0.5$, the interpolation is exactly the hyperbolic mean of these two images. We observe that in $\mathcal{W}^+$-space, both high-level attributes and low-level attributes changed together while in hyperbolic space, we can achieve more detailed editing on low-level attributes while keeping high-level attributes unchanged, while the radius can control the degree of change more precisely. When the hyperbolic radius is large, the category of the given image remains the same before reaching the middle point of the curve. This property allows us to generate diverse images of the given image without changing its category-relevant attributes. As the hyperbolic radius becomes smaller, the higher-level attributes gradually change in the early stage of the interpolation. The interpolation visualization on geodesic shows that the image gradually changes from fine-grained to abstract to fine-grained. These results also explain our method of adding details by rescaling after taking the geodesic which will lead to a relatively abstract average.

\section{Images Generated with Different Radii}
\label{Images Generated with Different Radii}
\begin{figure*}[!htb]
  \centering
   \includegraphics[width=0.8\linewidth]{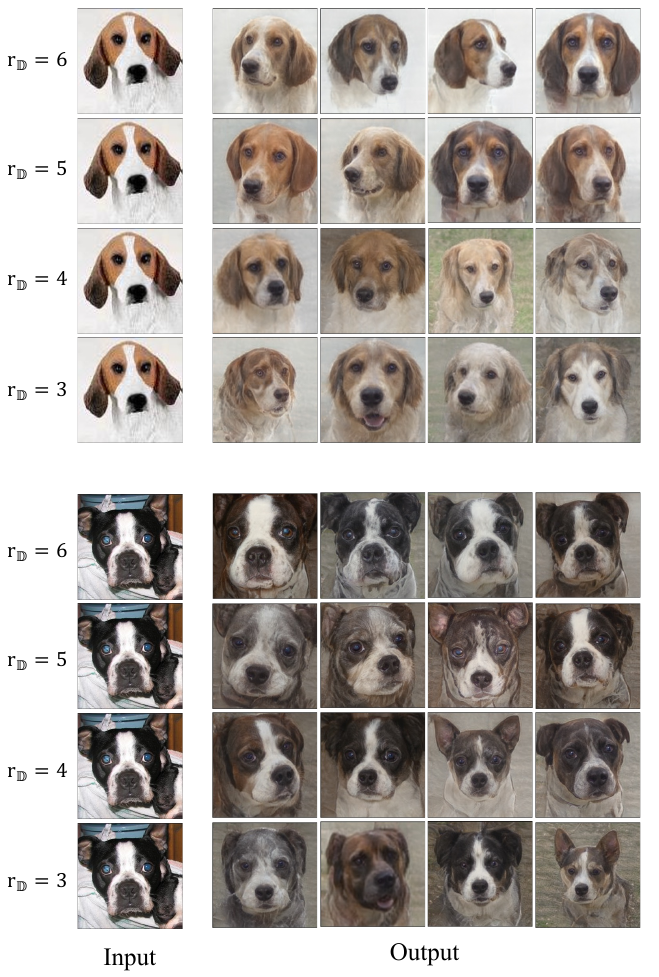}

   \caption{One-shot image generation by HAE on different radii on Animal Faces.}
   \label{fig:generation_radius}
\end{figure*}

In this section, we give more examples of images generated by HAE with different radii in the Poincaré disk. As \cref{fig:generation_radius} shows, the high-level attributes, {\it a.k.a.} category-relevant attributes do not change when the radius is large which allows us to generate diverse images without changing the category. However, the images generated by HAE become more abstract and semantically diverse when the hyperbolic radius $r_\mathbb{D}$ becomes smaller. The images gradually lose fine-grained details and change higher-level attributes as they move closer to the center of the Poincaré disk. For the few-shot image generation task, large radii work well since we want to change the category-irrelevant attributes of a given image. Nevertheless, our method HAE is not only capable of few-shot image generation but has great potential for other downstream tasks. For instance, HAE is able to generate a bunch of images of felines given an image of a cat. This can be done by rescaling the latent code to a relatively small radius in the hyperbolic space. Then we can add random perturbation to get the average code of multiple categories of felines. Finally, diverse images with fine-grained details of felines can be generated by moving those average codes to their children with larger radii.

\section{Comparison with State-of-the-art Few-shot Image Generation Method}
\label{Comparison with State-of-the-art few-shot image generation Method}
We also provide a comparison of images generated by the state-of-the-art method, \ie AGE~\cite{Ding22} and our methods on three datasets. As \cref{fig:Comparison} shows, our method is able to generate images with more semantic diversity. For instance, for dogs, HAE generates images with different light conditions and angles compared with images generated by AGE. Furthermore, for the woman in the third row from the bottom, HAE can change the image from a colored photo to a monochrome photo without changing her identity. However, our method also slightly changes some attributes compared with the original images, \eg, the hair color of dogs and the petal color of flowers. This is because the color varies within the category in these datasets which can indicate color is a category-irrelevant attribute for those categories. Therefore, our method does not change the category-relevant attributes but has more semantic diversity. Besides, images generated by HAE look more natural compared with images generated by AGE. Most importantly, AGE requires datasets with labels to learn the feature code book. However, the hierarchical representation in HAE can be learned using unsupervised or self-supervised learning if we have enough computing resources. Therefore, our work has great potential and can be applied to many other downstream tasks in future work.

\begin{figure}[h]
    \centering
    \includegraphics{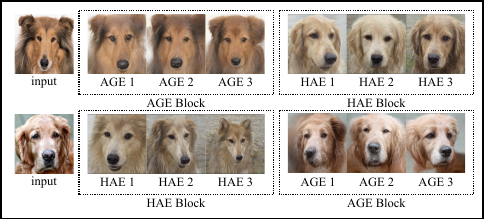}
    \caption{Illustration of shuffling in the user study, where inputs, blocks, and generated variants in each block were shuffled.}
    \label{fig:random_sampling}
\end{figure}

\section{User Study}
\label{User Study}
As mentioned, we conducted an extensive user study with a fully randomized survey. Results are shown in the main text. Specifically, we compared AGE and HAE in the following protocol:
\begin{enumerate}
    \item We randomly chose 20 images per dataset, and for each image, we then generated 3 variants using AGE and HAE, respectively. Overall, there were 60 original images and 180 generated variants in total.
    \item For each sample of each model, we grouped the 3 generated variants together, denoted as an \textit{image block}. We then shuffled the following orders in the dataset:
    1) order of images, 2) order of each block, 3) order of images in the block, an illustration is shown in \cref{fig:random_sampling}.
    \item We gathered 50 volunteers from various backgrounds who were asked to choose one image block for one sampled image based on their evaluation of image diversity and quality subjectively. The results are then re-arranged.
\end{enumerate}

\begin{figure*}[!htb]
  \centering
   \includegraphics[width=1\linewidth]{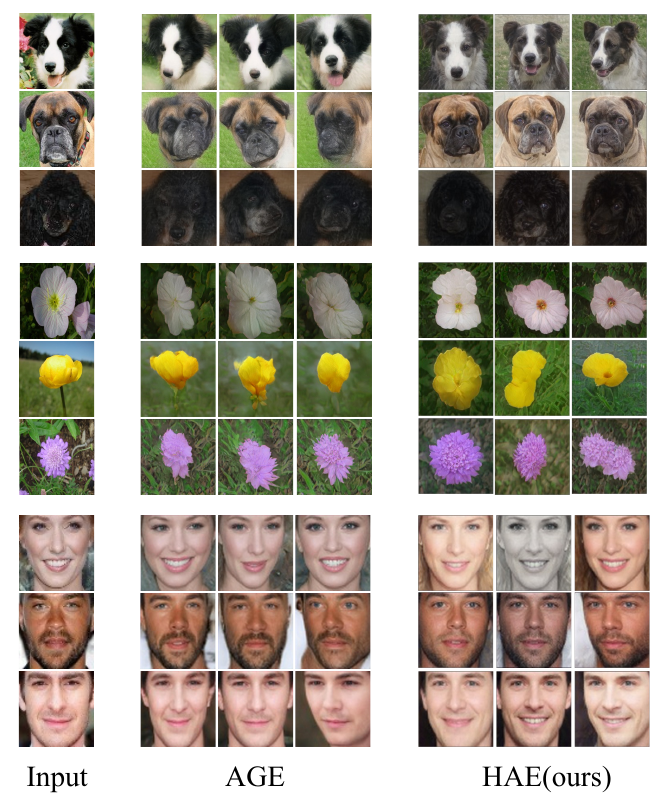}

   \caption{More examples used in the user study on three datasets.}
   \label{fig:UserStudy}
\end{figure*}

The result breakdowns are shown as follows: \textbf{Animal Faces}: 658/1000; \textbf{Flowers}: 523/1000; \textbf{VGGFaces}: 562/1000. We also provide more examples used in the user study in \cref{fig:UserStudy}.

\section{Additional Examples Generated by HAE}
\label{Additional Examples Generated by HAE}
\begin{figure*}[!htb]
  \centering
   \includegraphics[width=0.8\linewidth]{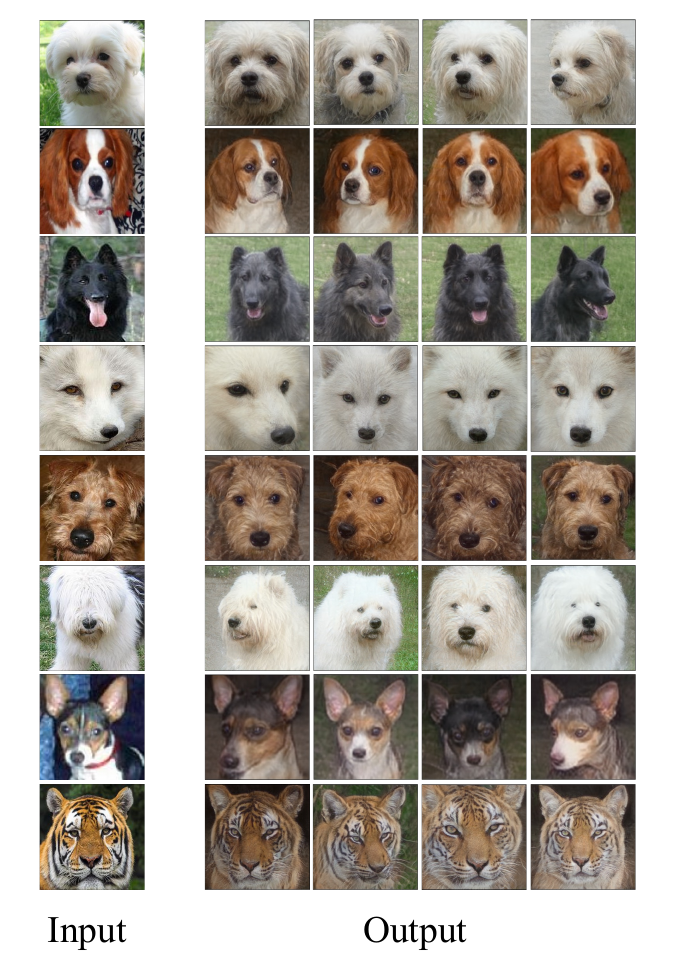}

   \caption{One-shot image generation by HAE on Animal Faces.}
   \label{fig:animals}
\end{figure*}

\begin{figure*}[!htb]
  \centering
   \includegraphics[width=0.8\linewidth]{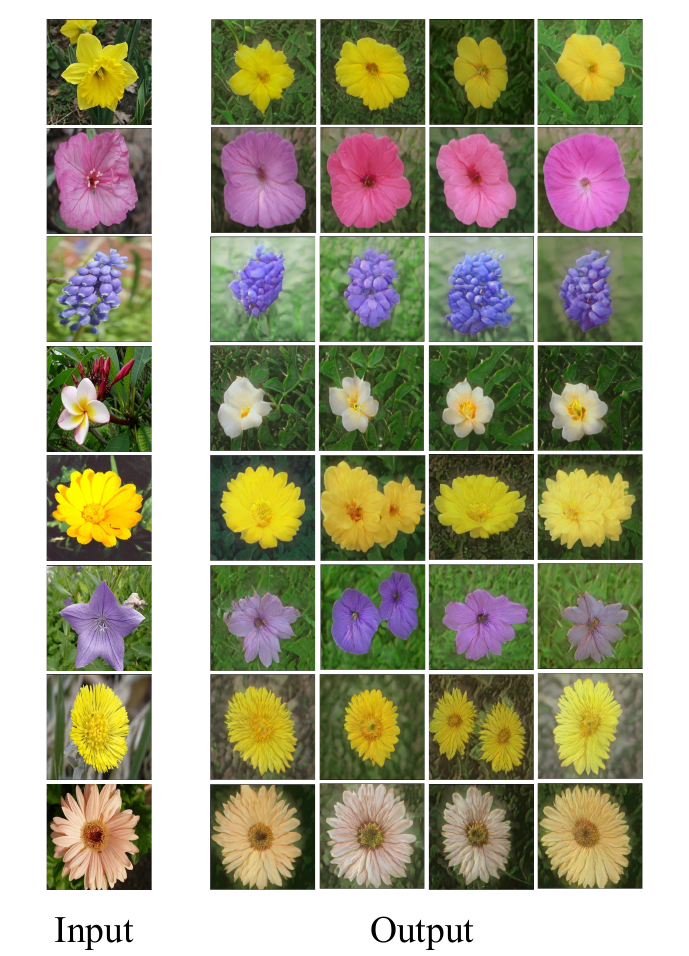}

   \caption{One-shot image generation by HAE on Flowers.}
   \label{fig:flowers}
\end{figure*}

\begin{figure*}[!htb]
  \centering
   \includegraphics[width=0.8\linewidth]{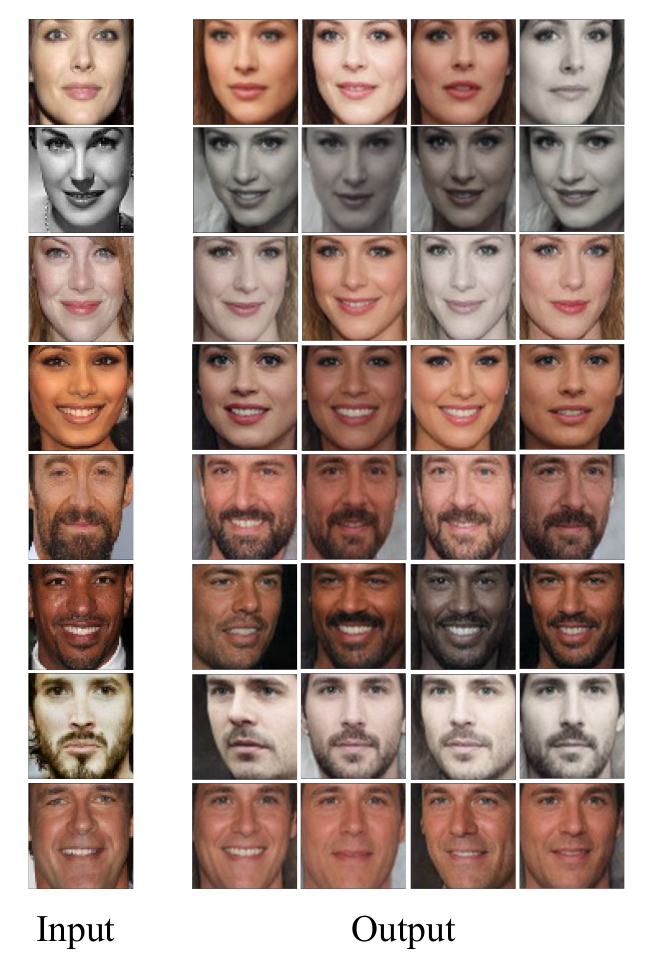}

   \caption{One-shot image generation by HAE on VGGFaces.}
   \label{fig:vggfaces}
\end{figure*}

We provide more samples generated by HAE in \cref{fig:animals}, \cref{fig:flowers}, and \cref{fig:vggfaces}. The radius we choose is 6.2126 and the length of perturbation is 8.

\begin{figure*}[!htb]
  \centering
   \includegraphics[width=1\linewidth]{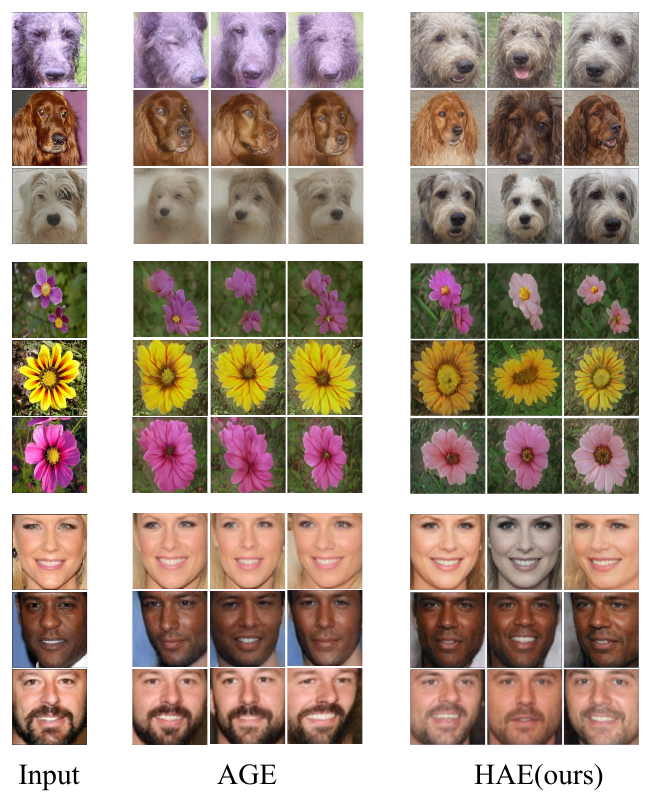}

   \caption{Comparison between images generated by AGE and HAE on Animal Faces, Flowers, and VGGFaces.}
   \label{fig:Comparison}
\end{figure*}

\end{appendix}

\end{document}